\documentclass[preprint,12pt]{elsarticle}

\usepackage{amssymb}
\usepackage{pifont}
\usepackage{amsmath}

\newcommand{\cmark}{\ding{51}}  % ✓
\usepackage{booktabs}
\usepackage{placeins}
\usepackage{xcolor}
\usepackage{xcolor,colortbl}

\usepackage{xcolor,colortbl}
\definecolor{axisblue}{RGB}{74,140,185}    % matches your figure’s blue
\definecolor{axisgreen}{RGB}{73,162,110}   % matches your figure’s green
\definecolor{axisorange}{RGB}{249,140,82}  % matches your figure’s orange

\usepackage{soul} % for highlighting

\usepackage{rotating}
\usepackage{pdflscape}   % for landscape pages
\usepackage{adjustbox}   % for resizing tables if needed
\usepackage{booktabs}    % for better horizontal rules
\usepackage{array}
\usepackage{threeparttable} % optional, if you want notes

\usepackage{supertabular} % better for multipage + captions than xtab
\usepackage{caption} % for manually placed captions

\newif\ifshownotes
\shownotestrue   % set to \shownotesfalse to hide everything
\ifshownotes
  
  \newcommand{\note}[1]{\textcolor{blue}{[\textbf{NOTE}: #1]}}
  
\else
  
  \newcommand{\note}[1]{}
\fi

\journal{Computer Speech \& Language}

\begin{document}

\begin{frontmatter}



\title{Which Evaluation for Which Model? A Taxonomy for Speech Model Assessment}


% \author{} %% Author name

% \affiliation{organization={},%Department and Organization

\author{Maureen de Seyssel\corref{cor1}}
\ead{mdeseyssel@apple.com}
\author{Eeshan Gunesh Dhekane}

\affiliation{Apple}
\cortext[cor1]{Corresponding author}

%% Abstract
\begin{abstract}
Speech foundation models have recently achieved remarkable capabilities across a wide range of tasks. However, their evaluation remains disjointed across tasks and model types. Different models excel at distinct aspects of speech processing and thus require different evaluation protocols. This paper proposes a unified taxonomy that addresses the question: \textit{Which evaluation is appropriate for which model?} 
The taxonomy defines three orthogonal axes: the \textbf{evaluation aspect} being measured, the \textbf{model capabilities} required to attempt the task, and the \textbf{task or protocol requirements} needed to perform it. We classify a broad set of existing evaluations and benchmarks along these axes, spanning areas such as representation learning, speech generation, and interactive dialogue. 
By mapping each evaluation to the capabilities a model exposes (e.g., speech generation, real-time processing) and to its methodological demands (e.g., fine-tuning data, human judgment), the taxonomy provides a principled framework for aligning models with suitable evaluation methods. It also reveals systematic gaps, such as limited coverage of prosody, interaction, or reasoning, that highlight priorities for future benchmark design. 
Overall, this work offers a conceptual foundation and practical guide for selecting, interpreting, and extending evaluations of speech models.
\end{abstract}

% \note{Include graphical abstract if necessary, but doesn't seem to be strictly required by Elsevier template}

% Not needed in source file as submitted separately
% \begin{highlights}
% \item Introduces a three-axis taxonomy for evaluation of speech foundation models
% \item Classifies evaluations by aspect, required model capabilities, and task protocol
% \item Maps existing evaluation tasks and benchmarks to the taxonomy for practical researcher guidance
% \item Reveals systematic gaps in current evaluation coverage

% \end{highlights}

%% Keywords
\begin{keyword}

Speech models \sep evaluation \sep benchmarking \sep taxonomy \sep speech processing



\end{keyword}

\end{frontmatter}




\section{Introduction}\label{sec:intro}
Modern speech-based models have rapidly expanded in capability and scope, spanning downstream applications such as automatic speech recognition and speech-to-speech translation, and intrinsic modeling abilities exemplified by spoken language models. As architectures diversify (e.g., encoder-only vs.\ encoder–decoder vs.\ generative) and modalities expand (speech-only, speech+text as a primary multimodal setting, and broader multimodal combinations such as speech–vision), evaluating these models has become a pressing challenge. A single benchmark or metric is rarely sufficient to capture the breadth of model behavior. For instance, a representation-focused model may be best tested on transcription or spoken language understanding tasks, whereas a generative model requires evaluation of output quality, intelligibility, naturalness, and content. This raises the central question of this paper: \emph{Which evaluation tasks are appropriate for which speech models?}

This question has motivated a wave of benchmark initiatives seeking to standardize evaluation across models and tasks.
Several benchmark suites have been proposed, including SUPERB~\cite{yang2024large}, SLUE~\cite{shon2022slue,shon2023slue}, and LeBenchmark~\cite{evain2021lebenchmark,parcollet2024lebenchmark}, which aggregate diverse downstream tasks spanning speech recognition, speaker identification, and emotion classification.
Complementary efforts such as the Zero Resource Speech Benchmark~\cite{dunbar2021zero}, Codec-SUPERB~\cite{wu2024codec}, and SALMon~\cite{maimon2025salmon} highlight complementary evaluation facets, from unsupervised representation learning to signal-level fidelity and acoustic consistency.
Yet, despite this diversity, most initiatives apply fixed task templates that do not always align with model capabilities. In practice, models differ substantially in their capabilities: some can generate speech, others cannot; some support real-time interaction, others do not. 
A more principled framework is therefore needed to align evaluations with model characteristics.

We use the term \emph{speech foundation models} to refer broadly to large speech or multimodal models trained on large-scale unlabeled or weakly supervised data and adaptable across multiple tasks. In line with the general definition of foundation models as systems trained on broad data at scale and adaptable to a wide range of downstream tasks~\cite{Bommasani2021FoundationModels}, this category encompasses both representation-oriented and generative models in the speech domain. It includes self-supervised speech representation models such as wav2vec 2.0~\cite{baevski2020wav2vec}, HuBERT~\cite{hsu2021hubert}, and WavLM~\cite{chen2022wavlm}, which learn general-purpose acoustic features from raw audio and are explicitly described as foundation models in the self-supervised speech literature~\cite{mohamed2022self}. It also includes the more recent spoken language models (SLMs)~\cite{arora2025landscape}: large generative or multimodal systems that model, interpret, and produce speech, often in combination with text, and that aim toward universal speech processing. Finally, this broader notion covers supervised or task-specific components that serve as building blocks for such systems, including acoustic encoders used in ASR, text-to-speech models, and speech-to-speech translation systems. We exclude fully fine-tuned task-specific models (e.g., end-to-end ASR) from this definition, focusing instead on the pre-trained encoders that form their foundation. Our goal is to provide an evaluation framework that applies uniformly across this full spectrum of speech foundation models, from self-supervised encoders to spoken language models and multimodal speech generation systems.

In this paper, we introduce a taxonomy for speech model evaluation that defines three orthogonal axes capturing complementary dimensions of the evaluation space:
(1) the \textbf{evaluation aspect} (what competency the task measures), 
(2) the \textbf{model capabilities required} (what the model must be able to do), and
(3) the \textbf{task or protocol requirements} (what the evaluation setup demands).
Rather than serving as an exhaustive survey, the taxonomy provides a structured framework for reasoning about evaluation choices.
Given a model with known capabilities and limitations, it enables researchers and practitioners to identify which evaluations are both feasible and meaningful.

\medskip
Our contributions are as follows:
\begin{itemize}
    % Minor: Should we skip "capability-aware" in this point? % Maureen : I'd actually keep it to emphasize the novelty of our taxonomy compared to usual surveys
    \item We introduce a capability-aware taxonomy of speech model evaluation, distinguishing evaluation aspects, model capabilities, and task or protocol requirements.
    \item We illustrate the taxonomy by categorizing a diverse selection of existing evaluation tasks and benchmarks (summarized in Tables~\ref{tab:tasks-catalogue}–\ref{tab:benchmarks}, \ref{sec:appendix-tables}).

    \item We provide practical guidance for choosing evaluations suited to different classes of models, illustrated with case studies and minimal recommended subsets.
\end{itemize}

This work complements prior surveys of audio and speech foundation models. For example, Mohamed et al.~\cite{mohamed2022self} review the development of self-supervised speech representation models that serve as foundation encoders for downstream tasks; Latif et al.~\cite{latif2023sparks} summarize recent advances in large audio models and benchmark efforts;
and Arora et al.~\cite{arora2025landscape} propose a taxonomy of spoken-language models along the axes of input/output modality and supervision. Our focus, in contrast, is on the \emph{evaluation} of such models. We provide a structured framework that connects model capabilities with feasible evaluation tasks, offering practical decision recipes for researchers and practitioners.
To the best of our knowledge, no prior work has proposed a unified, capability-aware framework for organizing evaluations across speech model types. Rather than aiming for exhaustiveness, our taxonomy is generative: it structures the evaluation space by what tasks measure, what models can do, and what each protocol requires, enabling the identification of missing aspects and guiding the design of future benchmarks.

The remainder of the paper is organized as follows.
Section~\ref{sec:related} reviews related work on existing benchmarks and prior taxonomy efforts.
Section~\ref{sec:taxonomy} presents our three-axis taxonomy of speech model evaluation and illustrates it with representative tasks and benchmarks.
Section~\ref{sec:guidance} provides practical guidance on selecting evaluations suited to different model types, while Section~\ref{sec:discussion} offers broader reflections and future directions.

To ground the discussion, we also include in Tables~\ref{tab:tasks-catalogue} and~\ref{tab:benchmarks} (\ref{sec:appendix-tables}) concise catalogues of representative evaluation tasks and benchmarks, with their references, descriptions, and associations.
These tables illustrate the diversity of current evaluation practices and serve as a reference framework that the subsequent taxonomy organizes along the three proposed axes.

\section{Related Work}\label{sec:related}

Evaluation practices for speech models have evolved rapidly alongside model capabilities.
Early work centered on broad benchmark suites such as SUPERB~\cite{yang2024large}, SLUE~\cite{shon2022slue,shon2023slue}, and LeBenchmark~\cite{evain2021lebenchmark,parcollet2024lebenchmark}, which standardized fine-tuning protocols across a variety of downstream tasks including phoneme recognition, speaker identification, and emotion classification.
These benchmarks played a central role in establishing reproducible comparisons for self-supervised encoders and remain widely used for measuring representation transferability.
Multilingual and language-specific variants later broadened coverage, including ML-SUPERB~\cite{shi2023ml} for multilingual speech understanding, FLEURS~\cite{conneau2023fleurs} and XTREME-S~\cite{conneau2022xtreme} for cross-lingual ASR and speech–text retrieval, and LeBenchmark~\cite{parcollet2024lebenchmark} for French evaluation.
While these suites offer valuable breadth, they share a similar structure: task-specific fine-tuning heads applied to frozen encoders, largely targeting representation-based rather than generative or interactive models.

As models evolved beyond recognition and classification applications, new benchmarks emerged to address generative and conversational abilities.
Codec-SUPERB~\cite{wu2024codec} extended the SUPERB framework to signal-level evaluation, covering speech resynthesis, enhancement, and codec reconstruction with metrics such as PESQ, STOI, and Mel distance.
The Zero Resource Speech Benchmark~\cite{versteegh2015zero,dunbar2017zero,dunbar2019zero,nguyen2020zero} instead focused on intrinsic, unsupervised evaluation, probing representation quality through tasks such as ABX discrimination and minimal-pair acceptability.
Other efforts targeted specific generative or interactive phenomena: MAD-Speech~\cite{futeral2025mad} quantified diversity in generated speech, SALMon~\cite{maimon2025salmon} measured acoustic-semantic consistency, EmphAssess~\cite{seyssel2024emphassess} and StressTest~\cite{yosha2025stresstest} tested prosodic emphasis transfer and understanding, and Full-Duplex-Bench~\cite{lin2025full} evaluated real-time turn-taking and backchanneling in spoken dialogue.
Together, these benchmarks illustrate the diversification of evaluation approaches but also highlight fragmentation: most apply fixed task templates that do not account for differences in model interfaces or capabilities.

In parallel, several surveys and taxonomies have attempted to synthesize this growing landscape.
Mohamed et al.~\cite{mohamed2022self} review the emergence of self-supervised speech representation models and their role as general-purpose encoders for downstream tasks.
Latif et al.~\cite{latif2023sparks} provide an overview of large audio models and associated benchmark trends.
Arora et al.~\cite{arora2025landscape} classify spoken language models along axes of input/output modality and supervision, framing the evolution from speech encoders to speech-aware large language models.
Cui et al.~\cite{cui2024recent} survey the architecture, training, and evaluation practices of spoken language models, emphasizing reasoning and instruction-following tasks.
Yang et al.~\cite{yang2025towards} propose a “holistic” taxonomy for evaluating large audio–language models (LALMs), organizing benchmarks into broad categories of auditory perception, reasoning, dialogue, and safety.
While these works provide valuable descriptive overviews, most are scoped to a specific subset of models, either speech encoders, spoken language models, or large audio–language models, and organize evaluations primarily by benchmark purpose.
None explicitly relate evaluations to the capabilities a model must expose or the procedural requirements each task entails.

\paragraph*{Our contribution}  
In contrast to these efforts, our work does not propose yet another benchmark suite, nor a survey of the existing ones. Instead, we introduce a capability-aware taxonomy of evaluation tasks themselves. By explicitly mapping evaluations to (i) the aspect being measured, (ii) the model capabilities required, and (iii) the task requirements, we provide a structured framework that helps researchers decide \emph{which evaluations are appropriate for a given model}. To our knowledge, this is the first systematic attempt to organize the evaluation space for speech foundation models along these lines.

\section{Taxonomy of Speech Model Evaluation}\label{sec:taxonomy}

\begin{figure}[htpb]
\centering
\includegraphics[width=0.9\linewidth]{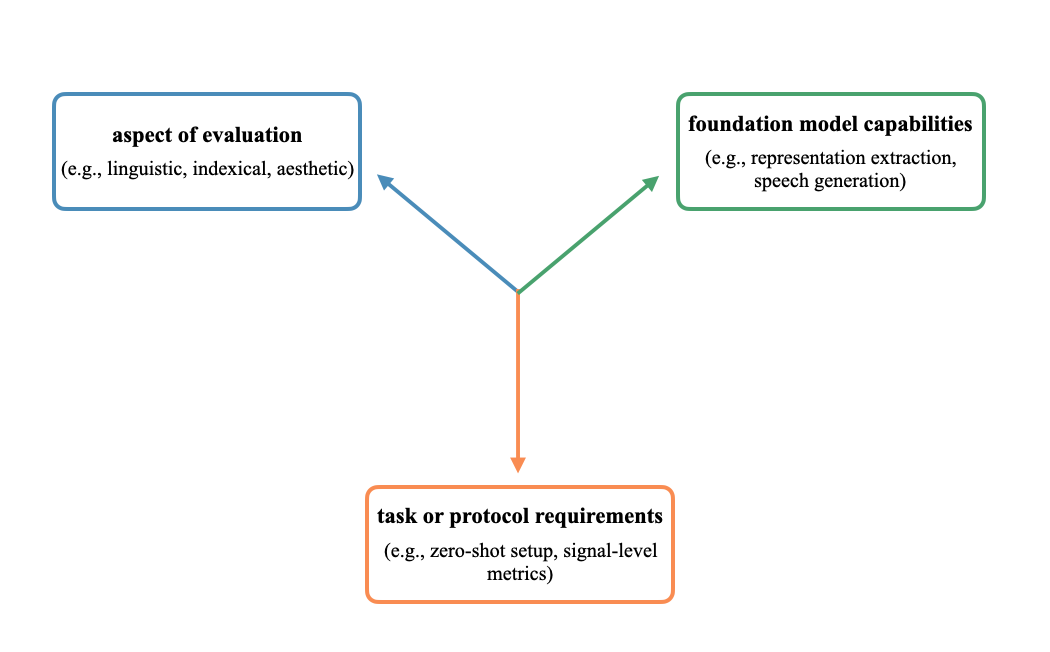}
\caption{The three orthogonal axes of the proposed taxonomy for speech model evaluation.}
\label{fig:taxonomy}
\end{figure}

Building on the gaps identified above, we now introduce a taxonomy that organizes the evaluation landscape for speech models. As summarized in Figure \ref{fig:taxonomy}, the taxonomy defines three orthogonal axes that together characterize any evaluation setting for a speech model:

\begin{enumerate}
    \item \colorbox{axisblue!20}{\textbf{Evaluation Aspect}}: What aspect of ability or knowledge does the evaluation measure?  
    \item \colorbox{axisgreen!20}{\textbf{Foundation Model Capability}}: What must the model be able to provide or support in order to perform the evaluation task?  
    \item \colorbox{axisorange!25}{\textbf{Task or Protocol Requirement}}: What does the evaluation setup demand in terms of data, fine-tuning, metrics, or human involvement?  
\end{enumerate}

Each axis is further divided into categories, as summarized in Table~\ref{tab:taxonomy-overview}.
While there is some overlap between the second and third axes, the distinction is deliberate:
the former captures what the \emph{model} must support (e.g., an ability to generate speech or produce representations),
whereas the latter captures what the \emph{evaluation procedure} requires (e.g., ground-truth labels, auxiliary models, or human raters), thus helping practitioners decide which evaluations are feasible and informative for their specific goals.
Maintaining both perspectives helps reveal mismatches between a model’s intrinsic capabilities and the external resources a task demands. For example, evaluation of downstream automatic speech recognition (ASR) requires that the model provide internal representations so that a decoder can map them to text (a \textbf{model requirement}),
while the evaluation itself may require fine-tuning on transcribed data or task-specific supervision (a \textbf{task requirement}).
Distinguishing these perspectives clarifies that a model may be theoretically capable of a task yet practically difficult to evaluate if the necessary resources are unavailable.

This framework offers a structured way to analyze both existing and emerging evaluations:
for any given task, one can specify (a) the primary aspect it measures, (b) the model capabilities required, and (c) the logistical constraints of the evaluation protocol.
Table~\ref{tab:taxonomy-overview} summarizes these three axes and their main subcategories, while extended tables in \ref{sec:appendix-tables} instantiate them with concrete benchmarks and tasks.

\begin{table}[t]
\caption{Summary of the three axes of the proposed taxonomy for speech model evaluation, with subcategories and short descriptions.}
\label{tab:taxonomy-overview}
\centering
\renewcommand{\arraystretch}{1.15}
\begin{adjustbox}{max width=\textwidth}
\begin{tabular}{p{8cm} p{11.5cm}}
\toprule
\textbf{Axis / Sub-Axis} & \textbf{Short Description} \\
\midrule
\rowcolor{axisblue!20}
\multicolumn{2}{l}{\textbf{\textit{Axis 1: Evaluation Aspect — What the evaluation measures}}} \\
Linguistic & Targets phonetic, lexical, or syntactic content of speech. \\
Understanding / Reasoning & Probes comprehension, inference, and world knowledge from spoken input. \\
Indexical & Focuses on stable identity traits such as speaker, accent, or language. \\
Paralinguistic & Evaluates expressive or prosodic cues (how something is said). \\
Acoustic / Signal Fidelity & Measures waveform fidelity, intelligibility, and objective signal quality. \\
Codec Reconstruction & Tests compression and resynthesis quality. \\
Aesthetic & Captures subjective naturalness, pleasantness, or style preference. \\
\midrule
\rowcolor{axisgreen!20}
\multicolumn{2}{l}{\textbf{\textit{Axis 2: Foundation Model Capability — What the model must support}}} \\
Speech Representation Extraction & The model provides latent features usable for analysis or downstream tasks. \\
% Plug-in Fine-Tuning & Allows lightweight supervised adaptation through task-specific training or added layers. \\
Discrete Unit Extraction & Outputs symbolic or quantized units derived from speech. \\
Sequence Probability Estimation & Assigns likelihoods or probabilities to sequences. \\
Speech Generation & Produces or reconstructs speech waveforms from textual, latent, or speech-based inputs. \\
Speech Continuation & Extends a spoken prompt with coherent new audio. \\
Multimodal (Speech + Text) & Processes both spoken and textual modalities. \\
Real-Time Interaction & Supports streaming, incremental inference, or conversational turn-taking. \\
\midrule
\rowcolor{axisorange!25}
\multicolumn{2}{l}{\textbf{\textit{Axis 3: Task / Protocol Requirement — What the evaluation setup demands}}} \\
Supervised Ground-Truth Data & Requires labeled references to compute evaluation metrics. \\
Fine-Tuning Required & Involves training on task data or supervised adaptation. \\
Zero-Shot / Direct Evaluation & Evaluates the model in its pre-trained form without adaptation. \\
Signal-Level Metrics & Relies on objective waveform-based quality scores. \\
Evaluation with Auxiliary Models & Uses external models (e.g., ASR, LLM) for scoring. \\
Human Evaluation & Depends on subjective judgments or preference ratings. \\
\bottomrule
\end{tabular}
\end{adjustbox}
\end{table}

Having outlined the three axes of the taxonomy, we now examine each in turn to clarify their scope and categories, starting with the evaluation aspects.

\FloatBarrier

\subsection{Evaluation Aspect Categories}\label{sec:aspect}

The \colorbox{axisblue!20}{\textbf{evaluation aspect}} axis of our taxonomy captures the primary property or capability that an evaluation task seeks to measure in a speech model.
As summarized under Axis~1 in Table~\ref{tab:taxonomy-overview}, this axis distinguishes between linguistic, understanding, indexical, paralinguistic, acoustic, codec, and aesthetic dimensions.
These categories are not always mutually exclusive. Each, however, highlights a distinct perspective on what an evaluation is probing and together cover the main levels of speech information: form, meaning, identity, expressivity, and signal quality.

\paragraph{\colorbox{axisblue!20}{Linguistic}}
While the term \emph{linguistic} could in principle encompass a broad range of communicative phenomena, we use it here in a narrower sense, covering the structural and symbolic content of speech: phonetic, lexical, and syntactic information. This choice follows long-standing distinctions in speech and language evaluation between form-level encoding and higher-level interpretation. Evaluating linguistic aspects is central because they capture the structural backbone of spoken communication: the mapping between acoustic form and symbolic meaning on which all higher-level understanding depends. Tasks in this group assess whether a model accurately encodes, decodes, or generates linguistic units such as phonemes, words, or syntactic structures. Common examples include downstream tasks such as phoneme recognition, keyword spotting, and automatic speech recognition (ASR). Beyond these, intrinsic evaluations like the ABX discrimination task~\cite{schatz2013evaluating,schatz2014evaluating} and the sWUGGY and sBLIMP~\cite{nguyen2020zero} minimal-pair tests probe whether a model’s representations distinguish minimal phonemic contrasts or grammatical versus ungrammatical sequences. 
Certain aspects of prosody, such as stress or phrase boundary marking when used to signal linguistic structure (e.g. prosAudit~\cite{de2023prosaudit}), are also included here. 
Recent work has extended text-based evaluation metrics to speech. For example, SpeechBERTScore, SpeechBLEU, and SpeechTokenDistance~\cite{saeki2024speechbertscore} adapt textual similarity metrics to discrete or embedding-based speech representations, while SpeechLMScore~\cite{maiti2023speechlmscore} uses a speech language model to assess the plausibility of generated outputs. Collectively, these metrics evaluate how closely the linguistic content of generated or recognized speech matches the intended message.

Finally, certain intrinsic unit-based evaluations, such as phoneme-normalized mutual information (PNMI) and phone/cluster purity~\cite{hsu2021hubert}, and VERT~\cite{lakhotia2021generative}, also probe how closely learned units align with phonemic categories. While we discuss them primarily under Acoustic / Signal Fidelity given their focus on representation fidelity, they can equally be viewed as linguistic evaluations insofar as they measure alignment with abstract phonemic structure.

\paragraph{\colorbox{axisblue!20}{Understanding and Reasoning}}
This aspect captures a model’s ability to connect spoken input to meaning and to reason about that meaning in context, whether through downstream tasks or intrinsic inference. 
Tasks in this category evaluate whether a model can interpret speech beyond its surface form and perform reasoning based on its semantic or pragmatic content.
This includes both models whose representations support downstream reasoning tasks (e.g., spoken question answering via textual decoding) and models that exhibit reasoning behaviour intrinsically, such as instruction-following or multimodal speech–language models prompted directly in speech.
The focus is therefore on whether the model captures meaning and pragmatic relations that extend beyond transcription, enabling inference, summarization, and contextually appropriate response generation.
Evaluations under this aspect vary in their setup and output modality: some rely on downstream supervised learning (e.g., spoken language understanding for intent classification or slot filling), others on generative modeling (e.g., spoken question answering or summarization), and others on probabilistic scoring of alternative continuations or answers.
Examples include SLU benchmarks~\cite{shon2022slue,shon2023slue}, story cloze tests~\cite{hassid2023textually}, spoken summarization~\cite{shon2023slue}, and MMLU-style reasoning from spoken prompts~\cite{sakshi2024mmau,cui2025voxeval,wang2025audiobench,yang2024air,chen2024voicebench}.
Together, these tasks span the continuum between representational support for semantic understanding and end-to-end reasoning from spoken input.

\paragraph{\colorbox{axisblue!20}{Indexical}}
This category concerns the identity of the speaker: who they are and what group they belong to, and therefore covers features such as accent and language.

Indexical information is a core property of speech: it reflects physiological and sociolinguistic factors that shape each speaker’s voice and enables recognition, attribution, and social inference. Tasks in this category assess whether a model captures stable identity markers such as voice, accent, dialect, or language. Examples include speaker identification~\cite{yang2024large,shor2020towards} and verification~\cite{yang2024large,parcollet2024lebenchmark}, language identification (LID)~\cite{shi2023ml,conneau2023fleurs,shor2020towards}, and accent diversity~\cite{futeral2025mad}. These evaluations differ from paralinguistic-oriented tasks (see below), which focus on temporary variations in expression (such as emotion or style), whereas indexical evaluations focus on stable traits that reveal who the speaker is or the community they belong to.

\paragraph{\colorbox{axisblue!20}{Paralinguistic}}
Beyond identity, speech conveys expressive and affective cues. This category covers evaluations of how something is said, rather than who is speaking. Paralinguistic cues convey attitude, affect, and interactional intent, making them essential for evaluating a model’s ability to reproduce the expressive and pragmatic dimensions of speech. Typical tasks include emotion recognition~\cite{yang2024large,shon2022slue,evain2021lebenchmark,shor2020towards,wu2024codec}, where models must classify the speaker’s affective state. Prosodic benchmarks such as EmphAssess~\cite{seyssel2024emphassess} similarly test whether models can preserve or transfer emphasis in appropriate semantic positions. Complementary work such as StressTest~\cite{yosha2025stresstest} extends this line of evaluation from emphasis transfer to prosody-based reasoning, introducing two tasks, Sentence Stress Detection and Sentence Stress Reasoning, that assess whether spoken language models can identify stressed words and infer speaker intent from prosodic emphasis. Paralinguistic information also plays a central role in instruction-following evaluations such as S2S Arena~\cite{jiang2025s2s}, which explicitly requires both understanding and generation of expressive cues. In addition, dialogue-oriented prosodic behaviors are included here, as captured by Full-Duplex-Bench~\cite{lin2025full}, which evaluates pause handling, backchanneling, smooth turn-taking, and user interruption.
Because prosody can serve multiple communicative functions, many of these evaluations intersect with other aspects such as linguistic structure (e.g., prosAudit~\cite{de2023prosaudit}) and understanding / reasoning (e.g. StressTest~\cite{yosha2025stresstest}).

\paragraph{\colorbox{axisblue!20}{Acoustic / Signal Fidelity}}
This category covers evaluations of low-level acoustic quality, whether intrinsic to a model’s outputs or relative to a reference signal. Acoustic fidelity underlies all other aspects of speech evaluation, as it determines whether a model preserves or reconstructs the physical properties of the signal that carry intelligibility and naturalness. A central resource here is Codec-SUPERB~\cite{wu2024codec}, which defines tasks for speech resynthesis, enhancement, and codec reconstruction, evaluated with standard signal-level metrics such as PESQ~\cite{rix2001perceptual}, STOI~\cite{taal2011algorithm}, SI-SNR, spectral distances (STFT, Mel), F0 correlation, and Mel-Cepstral Distortion (MCD). Source separation tasks are often measured by SDR improvement~\cite{vincent2006sdr}, while reference-free metrics such as the Fréchet Audio Distance (FAD)~\cite{kilgour2019frechet} compare embedding statistics between real and generated audio.

Certain intrinsic unit-based metrics, such as phoneme-normalized mutual information (PNMI), phone or cluster purity~\cite{hsu2021hubert},  and VERT~\cite{lakhotia2021generative}, probe the discriminative quality of learned tokenizations. While these metrics can be viewed as linguistic in nature, they are included here as measures of representation fidelity, since they quantify how well a model’s learned units preserve information from the input signal. Human-judgment-based assessments of naturalness and preference (e.g., mean opinion score) are covered separately under Aesthetics.

\paragraph{\colorbox{axisblue!20}{Aesthetics}}
Aesthetic evaluation captures the perceptual and affective response of human listeners, which is the ultimate measure of how natural, expressive, or pleasant generated speech sounds.
Complementing these objective metrics, aesthetic evaluations focus on subjective human judgments of audio quality and preference. Common tasks include Mean Opinion Score (MOS) estimation and pairwise preference tests, which remain the gold standard for assessing perceived naturalness and style. In addition, learned predictors have been proposed as scalable surrogates for human ratings, ranging from DNN-based MOS estimators \cite{mittag2019nisqa,reddy2021dnsmos} to recent aesthetic predictors such as Meta AudioBox Aesthetics \cite{tjandra2025meta}. While acoustic or signal-fidelity metrics focus on objective intelligibility or closeness to a reference waveform, aesthetic evaluations emphasize how outputs are experienced by listeners, qualities such as naturalness, expressivity, style, and overall pleasantness that cannot be fully captured by signal-based measures.

\medskip

Representative coverage of evaluation aspects across tasks is summarized in Table~\ref{tab:aspect-coverage} (\ref{sec:appendix-tables}), the first of three tables instantiating each axis of the taxonomy.
It is important to note that many evaluation tasks are inherently multi-faceted and can span several aspects at once. For instance, speech-to-speech translation involves both preservation of linguistic content and signal-level quality. Similar overlaps arise in prosody-related evaluations, which may serve either linguistic or paralinguistic functions. These intersections highlight that the categories in our taxonomy are not mutually exclusive but rather provide complementary perspectives on what a given evaluation measures.

%===========================

\subsection{Foundation Model Capability Requirements}\label{sec:capabilities}

Not every evaluation task can be practically applied to every model.  
Each task assumes that the model exposes certain abilities or interfaces, and a mismatch between task requirements and model capabilities can make an evaluation misleading or even impossible.  
We therefore identify a set of key \colorbox{axisgreen!20}{\textbf{foundation model capability requirements}}: basic interfaces that determine what kinds of evaluations a model can participate in.  
These can be read as a checklist of questions such as:  
\emph{Can the model expose frame-level representations? Can its features be adapted with lightweight probes? Can it produce discrete symbolic units? Can it generate speech outputs?}  

Although some requirements build on others (e.g., speech continuation presupposes speech generation), we describe them separately because they enable distinct evaluation setups. 
Each model may satisfy multiple capability criteria, depending on its architecture and training objectives.
The full set of capability types is summarized under Axis~2 in Table~\ref{tab:taxonomy-overview}, and the following subsections describe each in turn.

\paragraph{\colorbox{axisgreen!20}{Speech Representation Extraction}}
The model exposes internal speech representations (embeddings or latent features) that can be extracted and reused.
This is the most fundamental interface: without it, many evaluation tasks are impossible.
Representations may be used directly for intrinsic analyses of representational quality (e.g., ABX discrimination~\cite{schatz2013evaluating}, sWUGGY and sBLIMP~\cite{nguyen2020zero}, prosAudit~\cite{de2023prosaudit}),
or as inputs to downstream evaluations that train lightweight task-specific heads (see Axis~3 for protocol-level fine-tuning requirements).
Benchmarks such as SUPERB~\cite{yang2024large} assume this setup explicitly, as they require access to frozen representations that can be adapted under supervision. Most self-supervised speech encoders, such as wav2vec 2.0~\cite{baevski2020wav2vec} or HuBERT~\cite{hsu2021hubert}, provide frame-level or utterance-level features suitable for such evaluations.
In contrast, while encoder–decoder or sequence-to-sequence systems trained for end-to-end ASR or speech translation can in principle expose intermediate representations through their encoder outputs,
these are not always designed or documented for reuse in independent evaluations, which may limit their comparability in representation-based analyses.

\paragraph{\colorbox{axisgreen!20}{Discrete Unit Extraction}}
The model can output discrete token sequences from speech that function as symbolic units (e.g., phoneme-like or pseudo-text representations).
This capability underlies evaluations of unit quality, such as ABX discrimination,\footnote{While ABX discrimination can also be computed on continuous representations, it is often applied to discrete unit sequences to assess their phonemic consistency~\cite{nguyen2020zero}.} phoneme coverage, phoneme-normalized mutual information (PNMI)~\cite{hsu2021hubert}, or token diversity measures such as VERT~\cite{lakhotia2021generative}.
Benchmarks such as the Zero Resource Speech Benchmark~\cite{nguyen2020zero} include dedicated unit discovery tracks that apply these metrics to compare systems.
It is typically provided by models with an explicit discretization stage, such as vq-wav2vec~\cite{baevski2019vq}, HuBERT with clustered states~\cite{hsu2021hubert}, or related approaches that quantize hidden states into unit inventories.
Models that expose only continuous features cannot be evaluated directly on unit quality.
Codec models (e.g., EnCodec~\cite{defossez2022high}) also produce discrete tokens. However, since these are optimized for compression rather than symbolic abstraction, we exclude them from this category.

\paragraph{\colorbox{axisgreen!20}{Sequence Probability (Log-Likelihood) Estimation}}
The model can assign probabilities or likelihood scores to sequences of units, text, or audio tokens\footnote{
In most cases, this refers to the model being capable of computing \emph{normalized} sequence probabilities; for example, via standard log-likelihood or alternatively, via perplexities.
Even though this requirement directly holds for auto-regressive models, it does not necessarily exclude other model families, such as energy-based models and diffusion models.
This is because in the case of energy-based models, the unnormalized log-likelihood can be computed using the energy function as $\log\tilde{p}(\mathbf{x}) := -E_\theta(\mathbf{x})$. 
On the other hand, for diffusion models, it is proved that one can obtain estimation of the exact log-likelihood of data under the model~\cite{song2020score}, albeit with high variance. 
Thus, theoretically, one may still use the unnormalized version or a good estimation of the log-probability for ranking alternatives involved in the metric computation.
}. 

This goes beyond discrete unit extraction: instead of evaluating the quality of the units themselves, the requirement here is that the model defines a distribution over sequences, so that alternative candidates can be compared.  
Typical evaluations include perplexity on unit or phoneme sequences, minimal-pair acceptability tests such as sWUGGY (lexical) and sBLIMP (syntactic) \cite{nguyen2020zero}, as well as tasks where the model must select the more plausible continuation among alternatives, such as StoryCloze \cite{hassid2023textually}.  
Other benchmarks extend this format to broader reasoning and knowledge tests in the style of MMLU~\cite{hendrycksmeasuring}, where models must choose the correct option among several candidates.  
Models designed as autoregressive language-model-style generators, such as GSLM \cite{lakhotia2021generative}, pGSLM \cite{kharitonov2022text}, or SpiritLM \cite{nguyen2025spirit}, directly expose this capability. Masked-prediction models such as HuBERT~\cite{hsu2021hubert} do not define explicit sequence probabilities, although their outputs can in principle be used for approximate pseudo-likelihood evaluations when masked predictions are combined across positions. In contrast, embedding-only encoders or diffusion-based architectures generally cannot provide sequence probabilities and therefore cannot be evaluated in this way.

\paragraph{\colorbox{axisgreen!20}{Speech Generation}}
We use speech generation as an umbrella term covering any model capability to produce speech waveforms, whether through generative synthesis (creating new audio) or reconstructive processes (resynthesis, conversion, or decoding from latent representations).
This broad definition includes tasks such as end-to-end speech-to-speech translation, voice conversion, and resynthesis, as well as evaluations based on distributional audio metrics like the Fréchet Audio Distance (FAD)~\cite{kilgour2019frechet}.
Models without a speech decoder, such as text-only LLMs or encoder-only representation learners, cannot attempt such tasks.
Examples range from codec-based resynthesis systems such as EnCodec~\cite{defossez2022high} to unit-based or multimodal generators capable of producing speech from symbolic or latent inputs.
A stricter subcase of this requirement is speech continuation, in which the model must extend a spoken prompt with coherent new audio (discussed below).

\paragraph{\colorbox{axisgreen!20}{Speech Continuation}}
A specialized subset of speech generation is the ability to extend an existing spoken prompt with novel, contextually coherent audio, analogous to how text language models continue a text sequence.
Unlike reconstruction or transformation (e.g., codec resynthesis or voice conversion), continuation requires producing genuinely new content conditioned on prior speech, maintaining consistency in voice, prosody, and semantic flow.
This capability is essential for evaluating open-ended dialogue, conversational turn-taking, or long-form storytelling.
It is typically found in autoregressive or unit-based generative models with speech-in/speech-out capabilities (e.g., GSLM~\cite{lakhotia2021generative}, pGSLM~\cite{kharitonov2022text}, SpiritLM~\cite{nguyen2025spirit}) and in recent multimodal systems such as Qwen2-Audio~\cite{chu2024qwen2} and LLaMA-Omni~\cite{fang2024llama}.
Benchmarks such as SALMon~\cite{maimon2025salmon} and Full-Duplex-Bench~\cite{lin2025full} explicitly rely on this requirement to assess temporal and interactive coherence.

\paragraph{\colorbox{axisgreen!20}{Multimodal (Speech+Text) Capability}}
The model can process both speech and text, either as input or output. Many evaluation tasks rely on this cross-modal ability, for example in AudioBench~\cite{wang2025audiobench}, MMAU~\cite{sakshi2024mmau}, or AirBench~\cite{yang2024air}, where systems must handle spoken inputs alongside textual prompts or generate answers in different modalities.
Importantly, this differs from simple ASR-style setups in which a speech encoder is paired with an external text decoder: here, \textit{multimodality} refers to models that natively handle both modalities within a single architecture or pretraining objective.
Accordingly, only evaluations that require such joint processing are marked as depending on this capability.
Recent multimodal LLMs such as Qwen2-Audio~\cite{chu2024qwen2} and LLaMA-Omni~\cite{fang2024llama} exemplify this capability, supporting both speech and text interfaces within the same model.
This dimension is particularly relevant for spoken language models (SLMs), which explicitly unify speech perception, reasoning, and speech generation within a single framework.
Importantly, there is no requirement that training be based exclusively on speech; models pretrained primarily on text or multimodal data qualify as long as they can operate across modalities.

\paragraph{\colorbox{axisgreen!20}{Real-Time Streaming and Interaction}} The model can process speech incrementally with low latency and respond in a conversational setting. This capability is required for evaluations of real-time dialogue, such as handling interruptions, overlapping speech, or producing backchannels while listening. Models limited to offline processing can only handle fixed utterances, whereas interactive tasks demand architectures designed for streaming inference, such as transducers or blockwise/continuous processing. Benchmarks like Full-Duplex-Bench \cite{lin2025full} explicitly rely on this property, and have been used to evaluate systems such as dGSLM \cite{nguyen2023generative}, Moshi \cite{defossez2024moshi} or Salmonn-Omni \cite{yu2024salmonn}.

Together, these capabilities define what a model can expose, and thus, which evaluations are applicable.
They highlight how differences in model architecture, whether a system exposes representations, discrete units, or generative outputs, directly shape which tasks are feasible and which comparisons are valid.
Representative coverage of model capability requirements across evaluation tasks is provided in Table~\ref{tab:capreq-coverage} (\ref{sec:appendix-tables}), the second of three tables instantiating each axis of the taxonomy.

%=========================================================================

\subsection{Evaluation Task Requirements and Protocol}\label{sec:taskreq}

The third axis of our taxonomy describes what conducting the evaluation entails, namely, the \colorbox{axisorange!25}{\textbf{task requirements}}.
Whereas the previous axis specifies what a model must provide in order to be eligible for a task, this one captures what the practitioner chooses or assumes when setting up the evaluation.
Given a model’s capabilities (as defined along Axis 2), researchers can further filter which evaluations are feasible or appropriate depending on the level of supervision, data, or human involvement they are willing to rely on.
For instance, one may prefer zero-shot evaluation to test generalization without retraining, or fine-tuning to assess adaptability; similarly, one might choose automatic metrics for scalability or human ratings for subjective quality.

Although certain requirements depend on model capabilities (for instance, fine-tuning presupposes access to internal representations), it is useful to treat them separately; a model may technically support fine-tuning, yet the practitioner may choose not to use evaluations that require it. Likewise, the availability of ground-truth data or the need for human raters are properties of the evaluation setup rather than the model itself. As summarized under Axis 3 in Table~\ref{tab:taxonomy-overview}, this axis reflects the practical and methodological decisions that shape how an evaluation is conducted in practice.

While some requirements may naturally co-occur (such as fine-tuning and the use of ground-truth data), they refer to distinct aspects of the evaluation process: whether the model is adapted during the procedure, and what type of references or supervision the evaluation itself relies on. These distinctions are retained to make methodological assumptions transparent and to facilitate consistent comparison across studies.

\medskip
In practice, multiple requirements often co-occur within a single task. However, distinguishing them clarifies the trade-offs faced by researchers when selecting evaluations.
We therefore identify a set of common \textit{task requirements}, summarized below.
\paragraph{\colorbox{axisorange!25}{Supervised Ground Truth Data}} The evaluation requires labeled reference outputs in order to compute metrics. Most evaluation tasks fall under this category, with examples including ASR and speech translation, which use transcripts to compute word error rate or BLEU~\cite{papineni2002bleu}, and classification tasks, which need annotated examples to measure accuracy. This category also includes intrinsic evaluations such as ABX, sWUGGY, or sBLIMP, where correctness is defined with respect to labeled reference contrasts (e.g., phoneme identity, lexical validity, or grammaticality), even if the model itself is not fine-tuned on these labels. From a practitioner’s perspective, the appeal of supervised evaluations lies in their interpretability and comparability across models. However, they also depend on costly annotated data and may fail to generalize to domains or languages where such data are scarce. Some alternatives avoid this reliance altogether, such as intrinsic metrics (PNMI, cluster purity) or perceptual evaluations (MOS, preference tests), which can operate without labeled references.

\paragraph{\colorbox{axisorange!25}{Fine-Tuning Required}}
The evaluation involves training on task data, either by adapting the foundation model itself or by attaching and training one or more task-specific heads. This requirement presupposes that the model exposes trainable representations (as described under Axis 2). However, the decision to perform adaptation belongs to the evaluation setup rather than to the model itself. Many downstream evaluations fall into this category, such as classification tasks in SUPERB~\cite{yang2024large,shi2023ml}, SLUE~\cite{shon2022slue,shon2023slue}, or ParaLBench~\cite{zhang2024paralbench}.
Choosing such evaluations reflects an interest in adaptability: how well pre-trained representations support downstream learning under supervision. The trade-off is that fine-tuning introduces variability and resource costs, which can make it harder to isolate differences that stem from the foundation model rather than from training setup choices.

\paragraph{\colorbox{axisorange!25}{Zero-Shot or Direct Use (No Task-Specific Training)}} In these evaluations, the model is tested directly in its pre-trained form, without task-specific training. Examples include a generative model’s ability to follow instructions can be assessed by feeding prompts and judging the outputs, provided the model already supports generation. 
Representation-based tasks such as ABX discrimination \cite{schatz2013evaluating,schatz2014evaluating} can also be run on features from the frozen model without any adaptation or fine-tuning. 
Similarly, minimal-pair acceptability tests like sWUGGY or sBLIMP evaluate the probabilities a model assigns to sequences, again without training on task data. We mark tasks of this kind as “zero-shot,” in contrast to evaluations which require fine-tuning or additional task-specific training to be meaningful. 
% Minor: Used noun confounders.
Researchers may favor this setup when they wish to measure intrinsic competence, what the model can do without supervision, thus avoiding fine-tuning confounders, though such evaluations often require carefully designed prompts or metrics to remain fair across architectures. Distinguishing between fine-tuning and zero-shot setups also facilitates more principled selection of evaluations, allowing practitioners to filter tasks according to whether they wish to assess adaptability or intrinsic competence.

\paragraph{\colorbox{axisorange!25}{Signal-Level Metrics}}
Some evaluations rely on objective measures computed directly from the audio waveform rather than on symbolic annotations. Examples include speech quality and intelligibility metrics such as PESQ~\cite{rix2001perceptual}, STOI~\cite{taal2011algorithm}, and Mel-Cepstral Distortion (MCD), as well as distributional measures like Fréchet Audio Distance (FAD)~\cite{kilgour2019frechet}.   Learned quality predictors such as DNSMOS \cite{reddy2021dnsmos} do not comply within this category, as they rely on data-driven models rather than fixed signal-processing formulas.
Researchers may prefer these metrics when seeking a fully automatic and reproducible way to assess signal fidelity, how closely generated or reconstructed speech matches a reference, without requiring human raters or linguistic labels.
The requirement again lies not on the model but on the evaluation procedure; the evaluator must compute these metrics using specialized toolkits and, in many cases, paired reference recordings (e.g., clean vs.\ degraded speech).

\paragraph{\colorbox{axisorange!25}{Evaluation with Auxiliary Models}}
Certain evaluations instead rely on auxiliary models as scoring tools\footnote{We include here models used to generate symbolic or semantic judgments (e.g., ASR or LLM systems for automatic scoring), but not feature extractors employed purely within signal-level metrics such as FAD or DNSMOS.}. They offer scalable and automatic assessment, particularly useful for large generative systems or open-ended responses. A canonical example is using ASR to evaluate speech generation: tasks such as measuring WER on synthesized speech (e.g., ASR regeneration in Codec-SUPERB~\cite{wu2024codec}) rely on transcriptions produced by a pretrained ASR system. Similarly, benchmarks such as AudioBench~\cite{wang2025audiobench} and VoiceBench~\cite{chen2024voicebench} use automatic pipelines that combine ASR and large language models to assess the semantic accuracy or reasoning quality of generated responses. Such methods are efficient and scalable, making them attractive when human evaluation is impractical. 
However, they introduce an additional layer of bias and dependency; performance may reflect the auxiliary model’s strengths, limitations, or biases rather than the system under test, which must be carefully considered when interpreting results.

\paragraph{\colorbox{axisorange!25}{Human Evaluation}} Some aspects of speech quality or communicative effectiveness can only be judged by human listeners. Typical cases include naturalness and preference tests (e.g., MOS ratings or pairwise comparisons of synthesized speech). Benchmarks such as S2S Arena~\cite{jiang2025s2s} explicitly rely on human judgments to evaluate expressive and communicative adequacy in speech interactions. Researchers turn to human evaluations when perceptual quality or communicative appropriateness is central, dimensions that automatic metrics cannot yet capture reliably. The trade-off is cost and reproducibility: listener studies are time-consuming, and results may vary with participant demographics or experimental design.

\medskip

Table~\ref{tab:taskreq-coverage} (\ref{sec:appendix-tables}) summarizes how the evaluation tasks covered in our overview align with different protocol-level requirements.

\subsection{Mapping the Evaluation Landscape}\label{sec:tasks}

With the three axes in place, evaluation aspects, model capability requirements, and task protocol requirements, we now summarize how existing evaluation tasks for speech models align with this framework. The following overview provides a structured picture of the evaluation landscape, illustrating how the proposed dimensions distinguish between different types of assessments.

Tables~\ref{tab:tasks-catalogue}–\ref{tab:benchmarks} in \ref{sec:appendix-tables} instantiate the taxonomy through a set of illustrative summaries.
Table~\ref{tab:tasks-catalogue} lists representative evaluation tasks and benchmarks, grouped by their primary evaluation aspect (Section~\ref{sec:aspect});
% Minor: Typo.
Tables~\ref{tab:aspect-coverage}–\ref{tab:taskreq-coverage} then map each task respectively to the evaluation aspect, model capabilities required to attempt it, and to the practical requirements of running the evaluation.

Together, these tables provide a concise reference for researchers and practitioners. They make it possible to see at a glance which tasks apply to encoder-only, generative, multimodal, or interactive models, and where coverage gaps remain, for instance, the lack of standardized evaluations targeting full-duplex interaction or prosodic control.

Finally, Table~\ref{tab:benchmarks} (\ref{sec:appendix-tables}) consolidates the main benchmark suites discussed throughout the paper, showing how they relate to the proposed taxonomy in scope, coverage, and task types. This table illustrates how the proposed framework can organize existing benchmark suites and can be readily extended to accommodate future ones.

\section{Guidance: Choosing Evaluations for a Given Model}\label{sec:guidance}

The taxonomy presented in Section~\ref{sec:taxonomy} is not only descriptive but also prescriptive: it provides a framework for reasoning about which evaluations are meaningful for a given model type.
Its purpose is twofold. First, it serves as a practical guide for aligning model capabilities with compatible evaluation tasks, enabling practitioners to select feasible and informative tests, particularly through the third axis, which formalizes the methodological choices involved in conducting an evaluation (e.g., supervision, fine-tuning, human judgment).
Second, it exposes systematic gaps in the current evaluation landscape by highlighting capabilities for which no suitable benchmarks or metrics yet exist.
This section illustrates how to use the taxonomy in both ways.

\subsection{Using the Taxonomy to Select Evaluations}

This subsection illustrates how the taxonomy can be applied in practice through concrete examples.
The goal is to show how identifying a model’s exposed capabilities naturally constrains and guides the choice of meaningful evaluation tasks.
Each capability defines a set of tasks that are both technically feasible and theoretically well-grounded for that model, while ruling out others that would be misaligned or uninformative.

For instance, a model with \textit{speech generation} capability can be evaluated on waveform synthesis or resynthesis tasks such as text-to-speech, voice conversion, or reconstruction, which test its ability to produce natural and intelligible audio rather than to interpret input speech.
By contrast, a model that exposes \textit{internal representations} lends itself to intrinsic probing and representation-based evaluations such as ABX discrimination~\cite{schatz2013evaluating}, sWUGGY~\cite{nguyen2020zero}, or prosAudit~\cite{de2023prosaudit}, as well as frozen-feature downstream tasks included in suites such as SUPERB~\cite{yang2024large} or SLUE~\cite{shon2022slue,shon2023slue}.
This alignment between what a model exposes and what an evaluation measures underpins most current evaluation practice for self-supervised encoders: models like wav2vec 2.0 \cite{baevski2020wav2vec} and HuBERT \cite{hsu2021hubert} are typically assessed through representation-based tasks that test the structure and information content of their learned features rather than their ability to generate speech.

Similarly, models that output \textit{discrete tokens} enable evaluations targeting linguistic abstraction, for example PNMI, phone or cluster purity, and VERT~\cite{lakhotia2021generative}, as seen in quantized or clustered models such as vq-wav2vec~\cite{baevski2019vq} or HuBERT with clustered states~\cite{hsu2021hubert}.
Models that estimate \textit{sequence probabilities} allow probabilistic tests such as sWUGGY, sBLIMP, or StoryCloze~\cite{hassid2023textually}, while those designed for \textit{fine-tuning} are suited to supervised benchmarks like SUPERB, SLUE, or ParaLBench~\cite{zhang2024paralbench}.
Finally, models integrating \textit{multimodal processing}, \textit{streaming}, or \textit{speech continuation} capabilities invite evaluations targeting cross-modal reasoning (e.g.\ speech–text retrieval, spoken question answering), real-time dialogue (pause handling, backchanneling, user interruption), or open-ended speech generation (storytelling or conversational continuation).

Crucially, most evaluations depend on multiple capabilities at once.
For example, a prosody-aware speech generation task combines linguistic understanding with paralinguistic control, while interactive dialogue benchmarks add streaming and temporal coordination on top.
By cross-referencing capability requirements (Table~\ref{tab:capreq-coverage}) with protocol-level constraints (Table~\ref{tab:taskreq-coverage}), practitioners can identify which evaluations are jointly supported by a model’s interfaces and which would be conceptually or technically misaligned.
Again, these examples are illustrative rather than exhaustive, serving to demonstrate how the taxonomy can be used as a practical decision tool for selecting evaluations aligned with a model’s abilities.

The same reasoning can also be applied in reverse: given a model’s architecture and available interfaces, one can infer which evaluations are meaningful and which fall outside its scope.
At the opposite end, full-duplex dialogue systems integrate multiple capabilities, representation extraction, generation, multimodality, and streaming, and thus require composite evaluations covering comprehension, appropriateness, naturalness, and latency.

Beyond model capabilities, practitioners must also consider which aspect of evaluation they aim to probe, whether linguistic accuracy, expressive prosody, or perceptual quality.
A researcher interested in evaluating linguistic representation may prioritize tasks like ABX discrimination or sWUGGY, while one studying aesthetics or communicative naturalness may turn to MOS or human preference tests.
Conversely, if the goal is a broad assessment of general-purpose models, evaluations should ideally span multiple aspects of speech ability to capture both form- and meaning-related competence.

Together, these examples illustrate how the taxonomy supports principled evaluation design, ensuring that assessments remain both feasible and conceptually aligned with a model’s intended scope.

\subsection{Revealing Missing Evaluations}

Beyond guiding task selection, the taxonomy also exposes imbalances across evaluation aspects.  
Not all aspects of speech communication are equally represented in current evaluation practice, and some remain difficult to operationalize for certain model classes.  
Linguistic and acoustic fidelity are comparatively well covered: transcription accuracy, intelligibility, and signal quality have long-established metrics and benchmarks.  
By contrast, several other aspects remain underexplored.

\textit{Paralinguistic and prosodic aspects}, which refer to how something is said rather than what is said, are rarely evaluated systematically, particularly for generative and speech language models.  
While recent work introduces tests for emphasis transfer or emotional expressivity, no standardized metrics yet capture prosodic naturalness, pragmatic timing, or style variation at scale.  
Similarly, \textit{indexical aspects}, which concern speaker identity and sociolinguistic traits, are often treated as either nuisance factors or isolated classification tasks rather than as integral dimensions of communicative behaviour.  
Few evaluations measure whether generative models preserve or appropriately adapt such information in output speech.

\textit{Aesthetic and perceptual quality} is another underdeveloped area.  
Most generation benchmarks rely on mean opinion scores or a limited set of automatic predictors, with little cross-validation across domains or listener populations.  
Systematic methods for measuring subjective preference, perceived expressivity, or stylistic alignment remain largely absent.  
Finally, evaluations of \textit{understanding and reasoning} from spoken input are still dominated by text-based proxies, such as transcription accuracy or downstream NLP tasks, rather than speech-native tests of inference or pragmatic interpretation.

Taken together, these gaps reveal that current benchmarks disproportionately emphasize linguistic form and acoustic fidelity while leaving expressive, social, and pragmatic dimensions underrepresented.
The taxonomy thus serves not only as a map of existing evaluation coverage but also as a diagnostic tool, pinpointing which facets of speech communication are systematically neglected and where future evaluation efforts should focus.

\section{Discussion and Future Directions}\label{sec:discussion}

The taxonomy proposed in this paper is intended to provide conceptual structure in a rapidly evolving and fragmented evaluation landscape for speech models.  
Rather than introducing yet another benchmark or metric, it serves as a unifying reference framework that clarifies what each evaluation measures, what assumptions it makes about the model, and what it demands from the practitioner.  
By decomposing evaluations along three orthogonal axes: evaluation aspect, foundation model capability, and task protocol, the taxonomy enables systematic reasoning about model–task alignment.  
It helps determine not only which evaluations are feasible for a given system but also how to interpret results meaningfully in light of model design.

One immediate contribution of this framework is to make explicit the dependencies between model capabilities and evaluation validity.  
By explicitly specifying what interfaces a model provides (such as representations, discrete units, likelihoods, or speech generation) the taxonomy helps determine whether a task genuinely probes the model’s abilities or falls outside its scope.  
This makes it possible to avoid frequent mismatches, such as using signal-level metrics to assess encoder-only models or comparing systems that expose fundamentally different outputs.  
Equally, it supports more principled comparison across architectures: two models can only be meaningfully contrasted when they satisfy equivalent capability requirements and are evaluated under comparable task protocols.  
In this sense, the taxonomy provides not a static classification but a reasoning tool that links model properties, evaluation logic, and methodological design.

When applied to existing benchmarks, the taxonomy also reveals several systematic blind spots.  
Interactive and full-duplex systems, those requiring streaming, turn-taking, and real-time coordination, remain underserved by standardised metrics or datasets.  
Similarly, models producing discrete or latent semantic units lack established ways to assess meaning preservation or compositionality.  
Cross-modal and instruction-following systems are often tested with text-based metrics alone, leaving speech-grounded reasoning underexplored. Making such absences visible is one of the taxonomy’s strengths: it provides a clear map of where evaluation coverage is incomplete and where methodological innovation is most needed.

Although our focus is on evaluation frameworks applicable to speech foundation models, several application-specific domains have developed more specialized evaluation practices that could, in principle, be extended to this broader context.
For instance, while we note that current general-purpose benchmarks offer limited coverage of paralinguistic and expressive phenomena, such aspects are already addressed in more targeted applications like text-to-speech synthesis.
A recent example is EmergentTTS-Eval~\cite{manku2025emergenttts}, which introduces a large-scale benchmark for assessing prosodic and emotional expressiveness in TTS. These domain-focused efforts illustrate how advances from application-driven evaluation can inform future capability-aware frameworks for foundation models, particularly for the expressive and aesthetic dimensions of speech.

An important limitation of the present taxonomy is that it abstracts away from language-specific coverage.  
While the framework is defined in principle to be language-agnostic, focusing on evaluation \textit{tasks} rather than individual benchmarks, this abstraction can obscure a major practical gap: the overwhelming concentration of available resources in English.  
Most existing datasets, metrics, and even automatic evaluation pipelines are language-dependent, and their transfer to other languages is often non-trivial.  
As a result, the taxonomy does not by itself reveal which evaluations are linguistically portable or which aspects remain untested in multilingual settings.  
This limitation is particularly salient for representation learning and generative models trained across diverse linguistic domains, where the availability of benchmarks strongly constrains what can be meaningfully assessed.  
Extending the taxonomy to include a language coverage dimension would be an important next step toward a more globally representative evaluation landscape.

The taxonomy is, however, designed to evolve.  
As models acquire new capabilities (such as controllable prosody, affective intent, or multimodal grounding in visual and social context) new capability dimensions and evaluation aspects can be introduced while maintaining coherence within the existing framework.  
Future extensions may therefore address phenomena that current evaluation regimes overlook: conversational adaptivity, perceptual alignment between modalities, or social appropriateness in spoken interaction.  
By keeping the axes separable yet compatible, the taxonomy remains open to such additions without loss of interpretability.

Another promising direction for future work is to conduct large-scale comparative analyses of evaluation metrics across models and aspects.
Recent studies~\cite{minixhofer2025ttsds2} show that even widely used metrics can diverge in what they capture, underscoring that individual scores provide only a partial view of model behaviour. Building on the taxonomy proposed here, such analyses could systematically map which metrics correlate, overlap, or diverge across evaluation aspects and model types. This would not only clarify which measures are most informative for specific capabilities; it would also guide the design of more balanced and interpretable evaluation suites for speech foundation models.

In summary, this work contributes a unified, capability-aware framework that systematically links model characteristics, evaluation aspects, and methodological requirements, enabling principled comparison across diverse speech model types.
In the longer term, this structured view may also support a more unified evaluation ecosystem.  
Benchmarks and datasets could be indexed by capability requirements rather than by model family, allowing researchers to discover suitable evaluations automatically based on the interfaces a model provides.  
Researchers could adopt shared conventions for reporting model capabilities, making results more transparent and comparable across architectures and domains.
Ultimately, the value of this taxonomy lies not in fixing an exhaustive catalogue of tasks, but rather in providing a coherent conceptual map, one that makes evaluation choices explicit, exposes coverage gaps, and guides the design of future benchmarks toward greater interpretability and alignment.

\section{Conclusion}
As speech models continue to expand in scope, from encoding to generation, from perception to interaction, the need for consistent, interpretable, and theory-grounded evaluation becomes ever more critical.
The taxonomy proposed here offers a foundation for such consistency: a shared language for describing what evaluations test, what models enable, and how these align.
Importantly, the taxonomy is not intended as a fixed catalogue but as a living template: a structure designed to be continually filled and refined as new tasks, metrics, and benchmarks emerge. It provides a principled way to situate future evaluations within a coherent framework, ensuring that additions can be made without loss of interpretability. While new model architectures may eventually motivate additional categories, the current axes and subcategories should encompass the full range of existing evaluation practices.
By clarifying both the possibilities and the gaps, the taxonomy supports a more cumulative and reflective approach to speech model evaluation, one that can evolve in step with the models themselves and help the field move toward more comprehensive and interpretable assessment standards.

\section*{Acknowledgments}

The authors would like to thank Jie Chi and Natalie Schluter for their helpful feedback and discussions during the development and review of this work.

% \newpage

% \section*{Declaration of generative AI and AI-assisted technologies in the manuscript preparation process}

% During the preparation of this manuscript, the authors used an AI-assisted language tool to improve phrasing and formatting. The authors reviewed and edited the text to ensure accuracy and originality and take full responsibility for the content of the published article.

\appendix
% \appendix
\renewcommand{\thetable}{A.\arabic{table}}

% taxonomy_tables.tex
\clearpage
\section{Taxonomy Tables}
\label{sec:appendix-tables}

\noindent
This appendix compiles a set of illustrative tables that instantiate the three-axis taxonomy
introduced in Figure~\ref{fig:taxonomy}. 
Their goal is not to provide an exhaustive inventory (new evaluation tasks and benchmarks
continue to appear rapidly) but rather to demonstrate how the taxonomy can be applied and extended in practice.

\medskip
\noindent
Table~\ref{tab:tasks-catalogue} provides a concise catalogue of representative evaluation tasks
for speech models, with associated references, benchmark links, and short descriptions.
It serves as a reference list that the subsequent tables map onto the three axes of the taxonomy.

\medskip
\noindent
Tables~\ref{tab:aspect-coverage}–\ref{tab:taskreq-coverage} then align these tasks with the three
core axes of the framework:
(\emph{i}) the \textbf{aspect of evaluation} (the communicative or signal property being tested);
(\emph{ii}) the \textbf{foundation model capabilities} required to perform each evaluation; and
(\emph{iii}) the \textbf{task or protocol requirements} defining the evaluation setup.
A filled cell (\cmark) indicates that a task involves the corresponding dimension.

\medskip
\noindent
Finally, Table~\ref{tab:benchmarks} summarizes a subset of prominent benchmark suites that implement these evaluations,
highlighting their overall scope, language coverage, and included task types.
Together, these tables are intended as a \textit{living reference} illustrating how the proposed taxonomy
organizes a diverse and evolving evaluation landscape, and can be readily updated as new models and benchmarks emerge.

\medskip

%%%%%%%%%%%%%%%%%%%%%%%%%%%%%%%%%%%%%%%%%%%%%%%%%%%%%
% GENERAL SETTINGS FOR ALL TABLES
%%%%%%%%%%%%%%%%%%%%%%%%%%%%%%%%%%%%%%%%%%%%%%%%%%%%%
\scriptsize
\setlength{\tabcolsep}{3pt}
\renewcommand{\arraystretch}{1.15}

%%%%%%%%%%%%%%%%%%%%%%%%%%%%%%%%%%%%%%%%%%%%%%%%%%%%%
% TABLE A.1: EXTENDED TASK CATALOGUE — ORDER ALIGNED WITH TABLE 2
%%%%%%%%%%%%%%%%%%%%%%%%%%%%%%%%%%%%%%%%%%%%%%%%%%%%%
\clearpage
\begin{landscape}
\small
\setlength{\tabcolsep}{4pt}
\begin{center}

\begin{minipage}{1.05\linewidth}
\captionsetup{type=table}
\caption{Catalogue of evaluation tasks for speech models, ordered to match Table~\ref{tab:aspect-coverage}.}
\label{tab:tasks-catalogue}
\end{minipage}

\vspace{1em}

\tablefirsthead{
\toprule
\textbf{Evaluation Task} & \textbf{References} & \textbf{Related Benchmark(s)} & \textbf{Short Description} \\
\midrule}

\tablehead{
\multicolumn{4}{l}{\textbf{Table~\ref{tab:tasks-catalogue} (continued): Catalogue of evaluation tasks for speech models}}\\
\toprule
\textbf{Evaluation Task} & \textbf{References} & \textbf{Related Benchmark(s)} & \textbf{Short Description (continued)} \\
\midrule}

\tabletail{
\midrule
\multicolumn{4}{r}{\textit{Table continued on next page}} \\
\midrule}

\tablelasttail{\bottomrule}

\begin{supertabular}{p{4cm} p{2cm} p{3cm} p{11cm}}

Phoneme Recognition (PR) & \cite{yang2024large,shi2023ml} & SUPERB; ML-SUPERB &
Transcribes phoneme sequences from spoken audio; evaluates phonetic accuracy with Phone Error Rate (PER). \\

Automatic Speech Recognition (ASR) & \cite{yang2024large,shi2023ml,shon2022slue,conneau2023fleurs, conneau2022xtreme, evain2021lebenchmark}& SUPERB; ML-SUPERB; SLUE; FLEURS; XTREME-S; LeBenchmark; &
Converts speech to text; evaluates transcription accuracy via WER, CER or PER depending on language. \\
% Minor: Maybe we can also include edit-distance components (insertions, deletions, substitutions) that are also sometimes computed in standard packages.

ASR on Regenerated Speech (ASR\_regen) & \cite{wu2024codec} & Codec-SUPERB &
Tests speech resynthesis quality by running a pre-trained ASR model on generated audio to compute WER. \\

Out-of-Domain ASR (OOD-ASR) & \cite{yang2024large} & SUPERB &
Measures ASR performance on unseen or mismatched domains. \\

Keyword Spotting (KS) & \cite{yang2024large}  & SUPERB &
Detects predefined keywords in audio streams; evaluated with classification accuracy. \\

Query-by-Example (QbE) & \cite{yang2024large}  & SUPERB &
Searches for a spoken query within an audio database; uses Maximum Term-Weighted Value (MTWV) for evaluation. \\

Speaker Identification (SI) & \cite{yang2024large,shor2020towards} & SUPERB; NOSS &
Identifies which speaker produced an utterance; reports classification accuracy. \\

Speaker Verification (SV) & \cite{yang2024large, parcollet2024lebenchmark} & SUPERB; LeBenchmark 2.0 &
Verifies if two speech samples originate from the same speaker; evaluated with Equal Error Rate (EER). \\

SV on Regenerated Speech (SV\_regen) & \cite{wu2024codec} & Codec-SUPERB &
Tests preservation of speaker identity by applying a pre-trained SV model to generated audio. \\

Speaker Diarization (SD) & \cite{yang2024large} & SUPERB &
Segments multi-speaker recordings by speaker; evaluated with Diarization Error Rate (DER). \\

Language Identification (LID) & \cite{shi2023ml,conneau2023fleurs,shor2020towards} &
ML-SUPERB; FLEURS; NOSS &
Classifies spoken utterances by language; evaluated with accuracy. \\

Joint Multilingual ASR/LID task & \cite{shi2023ml}& ML-SUPERB &
Jointly predicts language ID and transcribes speech; computes PER/CER for ASR and accuracy for LID. \\

Emotion Recognition (ER) & \cite{yang2024large,shon2022slue, evain2021lebenchmark, shor2020towards} &
SUPERB; SLUE; LeBenchmark; NOSS &
Classifies emotional states from speech; evaluated by classification accuracy. \\

Emotion Recognition on Regenerated Speech (ER\_regen) & \cite{wu2024codec} & Codec-SUPERB &
Tests preservation of emotional expression using an external ER model on generated speech. \\

Audio Event Classification on Regenerated Speech (AEC\_regen) & \cite{wu2024codec} & Codec-SUPERB &
Assesses how well resynthesized audio preserves sound-event information via an external AEC model. \\

EmphAssess & \cite{seyssel2024emphassess} & EmphAssess &
Evaluates whether emphasis placement is preserved between input and regenerated speech. \\

Sentence Stress Detection (SSD) & \cite{yosha2025stresstest} & StressTest &
Tests whether a model can identify which word(s) in an utterance are stressed, given the ground-truth transcription; measures sensitivity to prosodic prominence using precision, recall, and F1. \\

Sentence Stress Reasoning (SSR) & \cite{yosha2025stresstest} & StressTest &
Evaluates whether a model can infer the speaker’s intended meaning from stress placement in speech; measured via accuracy using multiple-choice interpretation or LLM-as-judge scoring. \\

Intent Classification (IC) & \cite{yang2024large,conneau2022xtreme}; & SUPERB; XTREME-S &
Classifies user intent from spoken commands; measured by classification accuracy. \\

Slot Filling (SF) & \cite{yang2024large} & SUPERB &
Extracts semantic slots from spoken queries; evaluated with F1 and Character Error Rate (CER). \\

Speech Translation (ST) & \cite{yang2024large,conneau2022xtreme}& SUPERB; XTREME-S &
Translates speech in one language to text in another; evaluated with BLEU~\cite{papineni2002bleu}. \\

Voice Conversion (VC) & \cite{yang2024large} & SUPERB &
Converts one speaker’s voice into another’s while preserving linguistic content; evaluated via Mel-Cepstral Distortion (MCD). \\

Source Separation (SS) & \cite{yang2024large} & SUPERB &
Separates mixed-speaker audio into individual sources; evaluated with Signal-to-Distortion Ratio Improvement. \\

Speech Enhancement (SE) & \cite{yang2024large} & SUPERB &
Removes noise from speech; quality measured with PESQ and STOI. \\

ABX Discrimination & \cite{schatz2013evaluating,schatz2014evaluating,nguyen2020zero} &
ZeroSpeech Benchmark &
Compares distances between phonetic categories A and B to assess discriminability; lower error = better. \\

sWUGGY & \cite{nguyen2020zero} & ZeroSpeech Benchmark &
Evaluates lexical probability judgments by comparing model scores for real vs pseudo-words. \\

sBLIMP & \cite{nguyen2020zero} & ZeroSpeech Benchmark &
Tests syntactic acceptability: model assigns higher likelihood to grammatically correct sentences. \\

ProsAudit &  \cite{de2023prosaudit} &
ZeroSpeech Benchmark &
Assesses prosodic appropriateness by comparing likelihoods of prosodically well-formed vs ill-formed sentences. \\

Dialog Act Classification (DAC) & \cite{shon2023slue} & SLUE Phase-2 &
Classifies spoken utterances into dialogue act categories; evaluated via macro F1. \\

Question Answering (QA) & \cite{shon2023slue} & SLUE Phase-2 &
Finds spoken answer spans given spoken questions; evaluated via frame-level F1. \\

Speech Summarization (SUMM) & \cite{shon2023slue}& SLUE Phase-2 &
Generates text summaries from long speech inputs; evaluated with ROUGE~\cite{lin2004rouge}, METEOR~\cite{banerjee2005meteor}, and BERTScore~\cite{zhang2019bertscore}. \\
% Major: Do we need citations for these metrics?

Named Entity Localization (NEL) & \cite{shon2023slue} & SLUE Phase-2 &
Predicts time ranges of named entities; uses frame-F1 and word-F1. \\

Named Entity Recognition (NER) & \cite{shon2023slue} & SLUE Phase-2 &
Identifies and classifies named entities from speech; evaluated with micro F1. \\

Cross-modal Speech–Text Retrieval &  \cite{conneau2022xtreme,conneau2023fleurs} & FLEURS; XTREME-S &
Retrieves text from speech or vice versa using cross-modal embeddings; evaluated with Recall@k. \\

sSIMI & \cite{nguyen2020zero} & ZeroSpeech Benchmark &
Measures semantic similarity between spoken items. \\

Concept Prediction (CP) & \cite{evain2021lebenchmark} & LeBenchmark &
Predicts semantic concepts from speech; evaluated by Concept Error Rate (CER). \\

Automatic Speech-to-Text Translation (AST) & \cite{evain2021lebenchmark} & LeBenchmark &
Translates speech to target-language text; evaluated with BLEU. \\

Part-of-Speech Tagging (POS) & \cite{parcollet2024lebenchmark} & LeBenchmark 2.0 &
Classifies each spoken word’s part of speech; evaluated with F1. \\

Unlabeled Attachment Score (UAS) & \cite{parcollet2024lebenchmark} & LeBenchmark 2.0 &
Evaluates dependency parsing ignoring relation types. \\

Labeled Attachment Score (LAS) & \cite{parcollet2024lebenchmark} & LeBenchmark 2.0 &
Evaluates dependency parsing with labeled relations. \\

VERT (diVERsiTy score) & \cite{lakhotia2021generative} & — &
Measures trade-off between quality and diversity in generated speech using Auto-BLEU and Self-BLEU. \\

% Phoneme Language Coverage (PLC) & — & — &
% Compares phoneme distribution of generated vs reference language to assess phonetic coverage. \\

PNMI (Phoneme-Normalized Mutual Information) & \cite{hsu2021hubert} & — &
Quantifies alignment between discrete tokens and phonemes; higher PNMI indicates better clustering. \\

Phone / Cluster Purity & \cite{hsu2021hubert} & — &
Measures homogeneity of discovered phoneme clusters. \\

FAD (Fréchet Audio Distance) & \cite{kilgour2019frechet} & — &
Computes embedding-level audio fidelity between generated and reference speech using a pre-trained feature extractor (e.g., VGGish). The Fréchet inception distance between feature distributions yields the FAD score. \\
% Minor: Citation for VGGish. Precise reference to FID rather than the more generic Frechet distance.

MCD (Mel-Cepstral Distortion) & \cite{kubichek1993mel} & — &
Measures spectral distortion between original and regenerated speech using Mel-cepstral coefficients. \\

SI-SNR (Signal-to-Interference Ratio) & \cite{le2019sdr,luo2019conv} & — &
Quantifies signal quality; higher = less interference in regenerated speech. \\

STFT Distance & \cite{wu2024codec} & Codec-SUPERB &
Computes spectral distortion between original and resynthesized signals in STFT domain. \\

Mel Distance &\cite{wu2024codec} & Codec-SUPERB &
Computes distance between Mel-spectral representations of original and generated audio. \\

F0 Correlation (F0CORR) & \cite{wu2024codec} & Codec-SUPERB &
Measures correlation of fundamental frequency (F0) contours between original and regenerated speech. \\

STOI (Short-Time Objective Intelligibility) & \cite{wu2024codec} & Codec-SUPERB &
Measures intelligibility by correlating spectral features of original and regenerated speech. \\

PESQ (Perceptual Evaluation of Speech Quality) & \cite{wu2024codec} & Codec-SUPERB &
Assesses perceptual speech quality using auditory models;  correlates with human MOS. \\

POLQA & \cite{beerends2013perceptual} & — &
Objective perceptual speech-quality metric extending PESQ for wideband and VoIP conditions. \\

ViSQOL & \cite{hines2015visqol,chinen2020visqol} & — &
Spectro-temporal similarity metric producing MOS-LQO (Mean Opinion Score Listening Quality Objective). \\

SpeechBERT & \cite{saeki2024speechbertscore} & — &
Computes cosine similarity between feature embeddings of generated and reference speech to assess semantic alignment. \\

SpeechBLEU & \cite{saeki2024speechbertscore} & — &
Adapts BLEU to discrete speech tokens to measure n-gram overlap and phonetic alignment. \\

SpeechLMScore & \cite{maiti2023speechlmscore} & — &
Combines language modeling and acoustic modeling to assess generation quality and semantic accuracy. \\

SpeechTokenDistance & \cite{saeki2024speechbertscore} & — &
Measures edit distance or Jaro-Winkler distance between discrete speech token sequences. \\
% Minor: Do we add Jaro-Winkler distance citation?

Voice Diversity (MAD) & \cite{futeral2025mad} & MAD-Speech &
Measures diversity of speaker voices in generated speech via cosine dissimilarity or Vendi Score. \\
% Minor: Do we add Jaro-Winkler distance citation?

Gender Diversity & \cite{futeral2025mad}& MAD-Speech &
Assesses diversity of gender representation in generated speech embeddings. \\

Emotion Diversity & \cite{futeral2025mad} & MAD-Speech &
Measures variation in emotional states in generated speech. \\

Accent Diversity &\cite{futeral2025mad} & MAD-Speech &
Measures variation in accents in generated speech using accent-specific projection models. \\

Background Noise Diversity & \cite{futeral2025mad} & MAD-Speech &
Quantifies diversity of background noise types using noise-specific embedding models. \\

sStoryCloze / tStoryCloze &\cite{hassid2023textually} & — &
Evaluates story-continuation coherence by comparing likelihoods of coherent vs incoherent endings. \\

Multiple-Choice Questions (spoken QA) & \cite{cui2025voxeval} & VoxEval &
Tests reasoning by adapting MMLU into spoken multiple-choice format; evaluated via accuracy. \\

Generated Answer Analysis (spoken Q→text A) &  \cite{chen2024voicebench,wang2025audiobench} & VoiceBench; AudioBench &
Evaluates reasoning and answer quality when input is spoken but output text is analyzed via LLM-as-judge. \\

Multiple-Choice Question (text) & \cite{sakshi2024mmau,yang2024air,huang2024dynamic}  &
MMAU; Air-bench; Dynamic-SUPERB &
Text-based question on spoken input; evaluated with micro-averaged accuracy. \\

Open-Ended Questions (text) & \cite{yang2024air} & Air-bench &
Free-form spoken-input reasoning; scored with LLM-as-judge evaluation. \\

Acoustic Consistency (AC) & \cite{maimon2025salmon}  & SALMon &
Checks whether the model assigns higher likelihood to acoustically consistent vs inconsistent recordings. \\

Acoustic-Semantic Alignment (ASA) & \cite{maimon2025salmon}  & SALMon &
Measures alignment between acoustic and semantic attributes (e.g., background, sentiment). \\

Subjective MOS & \cite{international1996methods} & — &
Human-rated naturalness and intelligibility of generated speech; higher = better. \\

DNN-based MOS & \cite{reddy2021dnsmos} & — &
Predicts MOS using a DNN trained on human-rated pairs of original and regenerated speech. \\

NISQA & \cite{mittag2019nisqa}  & — &
Neural network–based non-intrusive speech-quality assessment model. \\

DNSMOS & \cite{reddy2021dnsmos}  & — &
Deep neural objective predictor of perceptual MOS for noisy speech. \\

Meta AudioBox Aesthetics & \cite{tjandra2025meta}  & Meta AudioBox &
Predicts aesthetic quality along Production Quality, Production Complexity, Content Enjoyment, and Content Usefulness axes. \\

ABX\_POS and ABX\_SEM & \cite{algayres2022dp}  & — &
Variants of ABX probing part-of-speech or semantic discrimination between token sequences. \\

S2S Arena &\cite{jiang2025s2s}  & S2S Arena &
Benchmark for evaluating speech-to-speech instruction following; includes pairwise human preference tests. \\

Pause Handling & \cite{lin2025full} & Full-Duplex-Bench &
Tests whether model avoids interrupting during user pauses; metric: Takeover Rate (TOR↓). \\

Backchanneling & \cite{lin2025full} & Full-Duplex-Bench &
Evaluates timing and appropriateness of short acknowledgements (e.g., “mm-hmm”); metrics: TOR, BC Frequency, JSD timing. \\

Smooth Turn-Taking & \cite{lin2025full}  & Full-Duplex-Bench &
Measures promptness and coordination in conversational turn-taking; latency and TOR metrics. \\

User Interruption Handling & \cite{lin2025full}  & Full-Duplex-Bench &
Assesses model response to user interruptions; TOR, GPT-4o relevance, and post-interruption latency. \\

% Other Downstream Classification Layers &  \cite{zhang2024paralbench}& ParaLBench &
% Collection of supervised downstream classification tasks probing model transferability. \\

\end{supertabular}
\end{center}
\end{landscape}

\clearpage

%%%%%%%%%%%%%%%%%%%%%%%%%%%%%%%%%%%%%%%%%%%%%%%%%%%%%
% TABLE 2: Evaluation Aspects
%%%%%%%%%%%%%%%%%%%%%%%%%%%%%%%%%%%%%%%%%%%%%%%%%%%%%

\clearpage
\begin{landscape}
\begin{center}

\captionsetup{type=table}
% \caption{Evaluation tasks and their coverage of \textit{Evaluation Aspects} (Axis~1 of the taxonomy). 
% A \cmark\ indicates that the task primarily or secondarily targets that aspect.}
\caption{\colorbox{axisblue!20}{Coverage of \textit{Evaluation Aspects} (Axis 1 of the taxonomy) for the representative tasks listed in Table~\ref{tab:tasks-catalogue}}. 
The table illustrates how each task maps onto the primary communicative or signal property it evaluates—linguistic, semantic, indexical, paralinguistic, acoustic, codec-related, or aesthetic. 
The listed tasks are not exhaustive of all existing evaluations, but correspond to the reference set used throughout this appendix, enabling direct cross-reading across Tables~\ref{tab:capreq-coverage}–\ref{tab:taskreq-coverage}. 
A filled cell (\cmark) indicates that the task primarily or secondarily targets the corresponding aspect.}

\label{tab:aspect-coverage}

\tablefirsthead{
\toprule
\textbf{Evaluation Task} &
\textbf{\shortstack{Linguistic}} &
\textbf{\shortstack{Understanding\\Reasoning}} &
\textbf{\shortstack{Indexical}} &
\textbf{\shortstack{Paralinguistic}} &
\textbf{\shortstack{Acoustic /\\Signal Fidelity}} &
\textbf{\shortstack{Codec\\Reconstruction}} &
\textbf{\shortstack{Aesthetic}} \\
\midrule
}

\tablehead{
\multicolumn{8}{l}{\colorbox{axisblue!20}{\textbf{Table~\ref{tab:aspect-coverage} (continued): Coverage of Evaluation Aspects (Axis~1 of the taxonomy)}}}\\

\toprule
\textbf{Evaluation Task} &
\textbf{\shortstack{Linguistic}} &
\textbf{\shortstack{Understanding\\Reasoning}} &
\textbf{\shortstack{Indexical}} &
\textbf{\shortstack{Paralinguistic}} &
\textbf{\shortstack{Acoustic /\\Signal Fidelity}} &
\textbf{\shortstack{Codec\\Reconstruction}} &
\textbf{\shortstack{Aesthetic}} \\
\midrule
}

\tabletail{
\midrule
\multicolumn{8}{r}{\textit{Table continued on next page}}\\
\midrule
}

\tablelasttail{\bottomrule}

\begin{supertabular}{p{5.5cm} *{7}{>{\centering\arraybackslash}p{2.2cm}}}

Phoneme Recognition (PR) & \cmark &  &  &  &  &  &  \\
Automatic Speech Recognition (ASR) & \cmark &  &  &  &  &  &  \\
ASR on regenerated speech (ASR\_regen) & \cmark &  &  &  & \cmark & \cmark &  \\
Out-of-Domain ASR (OOD-ASR) & \cmark &  &  &  &  &  &  \\
Keyword Spotting (KS) & \cmark &  &  &  &  &  &  \\
Query-by-Example (QbE) & \cmark &  &  &  &  &  &  \\
Speaker Identification (SI) &  &  & \cmark &  &  &  &  \\
Speaker Verification (SV) &  &  & \cmark &  &  &  &  \\
SV on regenerated speech (SV\_regen) &  &  & \cmark &  & \cmark & \cmark &  \\
Speaker Diarization (SD) &  &  & \cmark &  &  &  &  \\
Language Identification (LID) &  &  & \cmark &  &  &  &  \\
Joint Multilingual ASR/LID & \cmark &  & \cmark &  &  &  &  \\
Emotion Recognition (ER) &  &  &  & \cmark &  &  &  \\
ER on regenerated speech (ER\_regen) &  &  &  & \cmark & \cmark & \cmark &  \\
Audio Event Classification (AEC\_regen) &  &  &  &  & \cmark & \cmark &  \\
EmphAssess & \cmark &  &  & \cmark &  &  &  \\
Sentence Stress Detection (SSD) &  &  &  & \cmark &  &  &  \\
Sentence Stress Reasoning (SSR) &  & \cmark &  & \cmark &  &  &  \\
Intent Classification (IC) & \cmark & \cmark &  &  &  &  &  \\
Slot Filling (SF) & \cmark & \cmark &  &  &  &  &  \\
Speech Translation (ST) & \cmark & \cmark &  &  &  &  &  \\
Voice Conversion (VC) &  &  & \cmark &  & \cmark &  &  \\
Source Separation (SS) &  &  &  &  & \cmark & \cmark &  \\
Speech Enhancement (SE) &  &  &  &  & \cmark & \cmark &  \\
ABX & \cmark &  &  &  &  &  &  \\
sWUGGY & \cmark &  &  &  &  &  &  \\
sBLIMP & \cmark &  &  &  &  &  &  \\
ProsAudit & \cmark &  &  &  &  &  &  \\
Dialog Act Classification (DAC) &  & \cmark &  &  &  &  &  \\
Question Answering (QA) &  & \cmark &  &  &  &  &  \\
Summarization (SUMM) &  & \cmark &  &  &  &  &  \\
Named Entity Localization (NEL) &  & \cmark &  &  &  &  &  \\
Named Entity Recognition (NER) &  & \cmark &  &  &  &  &  \\
Cross-modal Speech–Text Retrieval &  & \cmark &  &  &  &  &  \\
sSIMI & \cmark & \cmark &  &  &  &  &  \\
Concept Prediction (CP) & \cmark & \cmark &  &  &  &  &  \\
Automatic Speech-to-Text Translation (AST) & \cmark & \cmark &  &  &  &  &  \\
Part-of-Speech Tagging (POS) & \cmark &  &  &  &  &  &  \\
Unlabeled Attachment Score (UAS) & \cmark &  &  &  &  &  &  \\
Labeled Attachment Score (LAS) & \cmark &  &  &  &  &  &  \\
VERT (diversity score) & \cmark &  &  &  & \cmark &  &  \\
% Phoneme Language Coverage (PLC) & \cmark &  &  &  & \cmark &  &  \\
Phoneme-Normalized Mutual Information (PNMI) & \cmark &  &  &  & \cmark &  &  \\
Cluster Purity & \cmark &  &  &  & \cmark &  &  \\
FAD &  &  &  &  & \cmark &  &  \\
MCD  &  &  &  &  & \cmark & \cmark &  \\
 SI-SNR  &  &  &  &  & \cmark & \cmark &  \\
STFT  &  &  &  &  & \cmark & \cmark &  \\
 MelDist  &  &  &  &  & \cmark & \cmark &  \\
 F0Corr &  &  &  &  & \cmark & \cmark &  \\

STOI  &  &  &  &  & \cmark & \cmark &  \\
 PESQ  &  &  &  &  & \cmark & \cmark &  \\
POLQA  &  &  &  &  & \cmark & \cmark &  \\
 ViSQOL &  &  &  &  & \cmark & \cmark &  \\

SpeechBERT  & \cmark &  &  &  &  &  &  \\
SpeechBLEU  & \cmark &  &  &  &  &  &  \\
 SpeechTokenDistance & \cmark &  &  &  & &  &  \\

SpeechLMScore & \cmark &  &  &  & &  &  \\
MAD (Voice, Gender, Emotion, Accent, Noise Diversity) &  &  & \cmark & \cmark & \cmark &  &  \\
sStoryCloze / tStoryCloze &  & \cmark &  &  &  &  &  \\
Multiple Choice (spoken Q + spoken A) &  & \cmark &  &  &  &  &  \\
Analysis of generated answer (spoken Q + text A) &  & \cmark &  &  &  &  &  \\
Multiple Choice (text Q) &  & \cmark &  &  &  &  &  \\
Open-Ended Questions (text) &  & \cmark &  &  &  &  &  \\
Acoustic Consistency (AC) &  &  &  &  & \cmark &  &  \\
Acoustic–Semantic Alignment (ASA) & \cmark & \cmark &  &  & \cmark &  &  \\
Subjective MOS &  &  &  &  & \cmark &  & \cmark \\
DNN-based MOS &  &  &  &  & \cmark &  & \cmark \\
NISQA  &  &  &  &  & \cmark &  & \cmark \\
Meta AudioBox Aesthetics &  &  &  &  &  &  & \cmark \\
ABX\_POS / ABX\_SEM & \cmark & \cmark &  &  &  &  &  \\
S2S Arena &  & \cmark &  & \cmark &  &  & \cmark \\
Pause Handling  &  &  &  & \cmark &  &  &  \\
Backchanneling &  &  &  & \cmark &  &  &  \\
 Turn Taking  &  &  &  & \cmark &  &  &  \\
Interruption &  &  &  & \cmark &  &  &  \\
\end{supertabular}
\end{center}
\end{landscape}

%%%%%%%%%%%%%%%%%%%%%%%%%%%%%%%%%%%%%%%%%%%%%%%%%%%%%
% TABLE 3: FOUNDATION MODEL CAPABILITIES (AXIS 2)
%%%%%%%%%%%%%%%%%%%%%%%%%%%%%%%%%%%%%%%%%%%%%%%%%%%%%

\clearpage
\begin{landscape}
\begin{center}

\captionsetup{type=table}
% \caption{Evaluation tasks and their required \textit{Foundation Model Capabilities} (Axis~2 of the taxonomy). 
% A \cmark\ indicates that a task explicitly depends on a given capability.}
\caption{\colorbox{axisgreen!20}{Evaluation tasks and their required \textit{Foundation Model Capabilities} (Axis~2 of the taxonomy).}
A \cmark\ indicates that a task explicitly depends on the corresponding capability.
The list is not exhaustive but covers all evaluation tasks discussed in Table~\ref{tab:tasks-catalogue},
allowing direct cross-reading across taxonomy axes.}

\label{tab:capreq-coverage}

\tablefirsthead{
\toprule
\textbf{Evaluation Task} &
\textbf{\shortstack{Speech\\Representation\\Extraction}} &
\textbf{\shortstack{Discrete\\Unit\\Extraction}} &
\textbf{\shortstack{Sequence\\Probability\\Estimation}} &
\textbf{\shortstack{Speech\\Generation }} &
\textbf{\shortstack{Speech\\Continuation }} &
\textbf{\shortstack{Multimodal\\(Speech+Text)\\Capability}} &
\textbf{\shortstack{Real-Time\\Interaction}} \\
\midrule
}

\tablehead{
\multicolumn{8}{l}{\colorbox{axisgreen!20}{\textbf{Table~\ref{tab:capreq-coverage} (continued): Evaluation tasks and their required foundation-model capabilities}}}\\
\toprule
\textbf{Evaluation Task} &
\textbf{\shortstack{Speech\\Representation\\Extraction}} &
\textbf{\shortstack{Discrete\\Unit\\Extraction}} &
\textbf{\shortstack{Sequence\\Probability\\Estimation}} &
\textbf{\shortstack{Speech\\Generation }} &
\textbf{\shortstack{Speech\\Continuation }} &
\textbf{\shortstack{Multimodal\\(Speech+Text)\\Capability}} &
\textbf{\shortstack{Real-Time\\Interaction}} \\
\midrule
}

\tabletail{
\midrule
\multicolumn{8}{r}{\textit{Table continued on next page}}\\
\midrule
}

\tablelasttail{\bottomrule}

\begin{supertabular}{@{}p{5.5cm} *{7}{>{\centering\arraybackslash}p{2.2cm}}@{}}

Phoneme Recognition (PR) & \cmark &  &  &  &  &  &  \\
Automatic Speech Recognition (ASR) & \cmark &  &  &  &  &  &  \\
ASR on regenerated speech (ASR\_regen) & \cmark &  &  & \cmark &  &  &  \\
Out-of-Domain ASR (OOD-ASR) & \cmark &  &  &  &  &  &  \\
Keyword Spotting (KS) & \cmark &  &  &  &  &  &  \\
Query-by-Example (QbE) & \cmark &  &  &  &  &  &  \\
Speaker Identification (SI) & \cmark &  &  &  &  &  &  \\
Speaker Verification (SV) & \cmark &  &  &  &  &  &  \\
SV on regenerated speech (SV\_regen) & \cmark &  &  & \cmark &  &  &  \\
Speaker Diarization (SD) & \cmark &  &  &  &  &  &  \\
Language Identification (LID) & \cmark &  &  &  &  &  &  \\
Joint Multilingual ASR/LID & \cmark &  &  &  &  &  &  \\
Emotion Recognition (ER) & \cmark &  &  &  &  &  &  \\
ER on regenerated speech (ER\_regen) & \cmark &  &  & \cmark &  &  &  \\
Audio Event Classification (AEC\_regen) & \cmark &  &  & \cmark &  &  &  \\
EmphAssess & \cmark &  &  & \cmark &  &  &  \\
Sentence Stress Detection (SSD) & \cmark &  &  &  &  & \cmark &  \\
Sentence Stress Reasoning (SSR) & \cmark &  &  &  &  & \cmark &  \\
Intent Classification (IC) & \cmark &  &  &  &  &  &  \\
Slot Filling (SF) & \cmark &  &  &  &  &  &  \\
Speech Translation (ST) & \cmark &  &  &  &  &  &  \\
Voice Conversion (VC) & \cmark &  &  & \cmark &  &  &  \\
Source Separation (SS) & \cmark &  &  & \cmark &  &  &  \\
Speech Enhancement (SE) & \cmark &  &  & \cmark &  &  &  \\
ABX & \cmark & \cmark &  &  &  &  &  \\
sWUGGY & \cmark & \cmark & \cmark &  &  &  &  \\
sBLIMP & \cmark & \cmark & \cmark &  &  &  &  \\
ProsAudit & \cmark & \cmark & \cmark &  &  &  &  \\
Dialog Act Classification (DAC) & \cmark &  &  &  &  & \cmark &  \\
Question Answering (QA) & \cmark &  &  &  &  &  &  \\
Summarization (SUMM) & \cmark &  &  &  &  & \cmark &  \\
Named Entity Localization (NEL) & \cmark &  &  &  &  &  &  \\
Named Entity Recognition (NER) & \cmark &  &  &  &  &  &  \\
Cross-modal Speech–Text Retrieval & \cmark &  &  &  &  & \cmark &  \\
sSIMI & \cmark & \cmark &  &  &  &  &  \\
Concept Prediction (CP) & \cmark &  &  &  &  &  &  \\
Automatic Speech-to-Text Translation (AST) & \cmark &  &  &  &  & \cmark &  \\
Part-of-Speech Tagging (POS) & \cmark &  &  &  &  &  &  \\
Unlabeled Attachment Score (UAS) & \cmark &  &  &  &  &  &  \\
Labeled Attachment Score (LAS) & \cmark &  &  &  &  &  &  \\
VERT (diversity score) &  &  &  & \cmark &  &  &  \\
% Phoneme Language Coverage (PLC) &  & \cmark &  & \cmark &  &  &  \\
Phoneme-Normalized Mutual Information (PNMI) &  & \cmark &  &  &  &  &  \\
Cluster Purity & \cmark & \cmark &  &  &  &  &  \\
FAD &  &  &  & \cmark &  &  &  \\
MCD &  &  &  & \cmark &  &  &  \\
 SI-SNR  &  &  &  & \cmark &  &  &  \\
STFT &  &  &  & \cmark &  &  &  \\
MelDist &  &  &  & \cmark &  &  &  \\
 F0Corr &  &  &  & \cmark &  &  &  \\
STOI &  &  &  & \cmark &  &  &  \\
PESQ &  &  &  & \cmark &  &  &  \\
POLQA &  &  &  & \cmark &  &  &  \\
ViSQOL &  &  &  & \cmark &  &  &  \\
SpeechBERT & \cmark &  &  & \cmark &  &  &  \\
SpeechBLEU & \cmark &  &  & \cmark &  &  &  \\
SpeechTokenDistance & \cmark &  &  & \cmark &  &  &  \\
SpeechLMScore & \cmark &  &  & \cmark &  &  &  \\

MAD (Voice, Gender, Emotion, Accent, Noise Diversity) & \cmark &  &  & \cmark &  &  &  \\
sStoryCloze / tStoryCloze & \cmark & \cmark & \cmark &  &  &  &  \\
Multiple Choice (spoken Q + spoken A) & \cmark &  &  &  &  & \cmark &  \\
Analysis of generated answer (spoken Q + text A) & \cmark &  &  & \cmark &  & \cmark & \cmark \\
Multiple Choice (text Q) &  &  &  &  &  & \cmark &  \\
Open-Ended Questions (text) &  &  &  &  &  & \cmark & \cmark \\
Acoustic Consistency (AC) & \cmark &  &  & \cmark &  &  &  \\
Acoustic–Semantic Alignment (ASA) & \cmark &  & \cmark & \cmark &  & \cmark &  \\
DNN-based MOS & &  &  & \cmark &  &  &  \\
Subjective MOS &  &  &  & \cmark &  &  & \cmark \\
NISQA  &  &  &  & \cmark &  &  & \cmark \\
Meta AudioBox Aesthetics &  &  &  & \cmark & \cmark &  &  \\
ABX\_POS / ABX\_SEM & \cmark & \cmark &  &  &  &  &  \\
S2S Arena & \cmark &  &  & \cmark & \cmark & \cmark & \cmark \\
Pause Handling  & \cmark &  &  &  & \cmark &  & \cmark \\
 Backchanneling & \cmark &  &  &  & \cmark &  & \cmark \\

Turn Taking & \cmark &  &  &  & \cmark &  & \cmark \\

Interruption & \cmark &  &  &  & \cmark &  & \cmark \\

\end{supertabular}
\end{center}
\end{landscape}

\clearpage
\begin{landscape}
\begin{center}

\captionsetup{type=table}
\caption{\colorbox{axisorange!25}{Evaluation tasks and their required \textit{Task / Protocol Requirements} (Axis~3 of the taxonomy).}
A \cmark\ indicates that a task explicitly depends on the corresponding requirement.
The list is not exhaustive but covers all evaluation tasks discussed in Table~\ref{tab:tasks-catalogue},
allowing direct cross-reading across taxonomy axes.}
\label{tab:taskreq-coverage}

\tablefirsthead{
\toprule
\textbf{Evaluation Task} &
\textbf{\shortstack{Supervised\\Ground-Truth\\Data}} &
\textbf{\shortstack{Fine-Tuning\\Required}} &
\textbf{\shortstack{Zero-Shot/\\Direct\\Evaluation}} &
\textbf{\shortstack{Evaluation\\with\\Auxiliary\\Models}} &
\textbf{\shortstack{Signal-Level\\Metric}} &
\textbf{\shortstack{Human\\Evaluation}} \\
\midrule
}

\tablehead{
\multicolumn{7}{l}{\colorbox{axisorange!25}{\textbf{Table~\ref{tab:taskreq-coverage} (continued): Evaluation tasks and their required task / protocol conditions}}}\\
\toprule
\textbf{Evaluation Task} &
\textbf{\shortstack{Supervised\\Ground-Truth\\Data}} &
\textbf{\shortstack{Fine-Tuning\\Required}} &
\textbf{\shortstack{Zero-Shot/\\Direct\\Evaluation}} &
\textbf{\shortstack{Evaluation\\with\\Auxiliary\\Models}} &
\textbf{\shortstack{Signal-Level\\Metric}} &
\textbf{\shortstack{Human\\Evaluation}} \\
\midrule
}

\tabletail{
\midrule
\multicolumn{7}{r}{\textit{Table continued on next page}}\\
\midrule
}

\tablelasttail{\bottomrule}

\begin{supertabular}{@{}p{5.5cm} *{6}{>{\centering\arraybackslash}p{2.3cm}}@{}}

Phoneme Recognition (PR) & \cmark & \cmark &  &  &  &  \\
Automatic Speech Recognition (ASR) & \cmark & \cmark &  &  &  &  \\
ASR on regenerated speech (ASR\_regen) & \cmark &  &  & \cmark & &  \\
Out-of-Domain ASR (OOD-ASR) & \cmark & \cmark &  &  &  &  \\
Keyword Spotting (KS) & \cmark & \cmark &  &  &  &  \\
Query-by-Example (QbE) & \cmark & \cmark &  &  &  &  \\
Speaker Identification (SI) & \cmark & \cmark &  &  &  &  \\
Speaker Verification (SV) & \cmark & \cmark &  &  &  &  \\
SV on regenerated speech (SV\_regen) &  &  &  & \cmark &  &  \\
Speaker Diarization (SD) & \cmark &  &  &  & \cmark &  \\
Language Identification (LID) & \cmark & \cmark &  &  &  &  \\
Joint Multilingual ASR/LID & \cmark & \cmark &  &  &  &  \\
Emotion Recognition (ER) & \cmark & \cmark &  &  &  &  \\
ER on regenerated speech (ER\_regen) & \cmark &  &  & \cmark &  &  \\
Audio Event Classification (AEC\_regen) & \cmark &  &  & \cmark &  &  \\
EmphAssess & \cmark &  &  & \cmark &  &  \\
Sentence Stress Detection (SSD) & \cmark &  & \cmark & \cmark &  &  \\
Sentence Stress Reasoning (SSR) & \cmark &  & \cmark & \cmark &  &  \\

Intent Classification (IC) & \cmark & \cmark &  &  &  &  \\
Slot Filling (SF) & \cmark & \cmark &  &  &  &  \\
Speech Translation (ST) & \cmark & \cmark &  &  &  &  \\
Voice Conversion (VC) & \cmark &  &  & \cmark &  &  \\
Source Separation (SS) & \cmark &  &  &  & \cmark &  \\
Speech Enhancement (SE) & \cmark &  &  &  & \cmark &  \\
ABX & \cmark &  & \cmark &  &  &  \\
sWUGGY & \cmark &  & \cmark &  &  &  \\
sBLIMP & \cmark &  & \cmark &  &  &  \\
ProsAudit & \cmark &  & \cmark &  &  &  \\
Dialog Act Classification (DAC) & \cmark & \cmark &  &  &  &  \\
Question Answering (QA) & \cmark & \cmark &  &  &  &  \\
Summarization (SUMM) & \cmark & \cmark &  &  &  &  \\
Named Entity Localization (NEL) & \cmark & \cmark &  &  &  &  \\
Named Entity Recognition (NER) & \cmark & \cmark &  &  &  &  \\
Cross-modal Speech–Text Retrieval & \cmark &  & \cmark &  &  &  \\
sSIMI & \cmark &  & \cmark &  &  &  \\
Concept Prediction (CP) & \cmark & \cmark &  &  &  &  \\
Automatic Speech-to-Text Translation (AST) & \cmark & \cmark &  &  &  &  \\
Part-of-Speech Tagging (POS) & \cmark & \cmark &  &  &  &  \\
Unlabeled Attachment Score (UAS) & \cmark & \cmark &  &  &  &  \\
Labeled Attachment Score (LAS) & \cmark & \cmark &  &  &  &  \\
VERT (diversity score) &  &  & \cmark &  &  &  \\
% Phoneme Language Coverage (PLC) &  &  & \cmark &  &  &  \\
Phoneme-Normalized Mutual Information (PNMI) & \cmark &  & \cmark &  &  &  \\
Cluster Purity & \cmark &  & \cmark &  &  &  \\
FAD &  &  &  & \cmark & \cmark &  \\
MCD &  &  &  &  & \cmark &  \\
SI-SNR &  &  &  &  & \cmark &  \\
STFT &  &  &  &  & \cmark &  \\
MelDist &  &  &  &  & \cmark &  \\
F0Corr &  &  &  &  & \cmark &  \\
STOI &  &  &  &  & \cmark &  \\
PESQ &  &  &  &  & \cmark &  \\
POLQA &  &  &  &  & \cmark &  \\
ViSQOL &  &  &  &  & \cmark &  \\

SpeechBERT & \cmark &  & \cmark & \cmark & \cmark &  \\
SpeechBLEU & \cmark &  & \cmark &  & \cmark &  \\
SpeechTokenDistance & \cmark &  & \cmark &  & \cmark &  \\
SpeechLMScore &  &  & \cmark & \cmark &  &  \\
MAD (Voice, Gender, Emotion, Accent, Noise Diversity) & \cmark &  & \cmark & \cmark &  &  \\
sStoryCloze / tStoryCloze &  \cmark &  & \cmark &  &  &  \\
Multiple Choice (spoken Q + spoken A) & \cmark &  & \cmark &  &  &  \\
Analysis of generated answer (spoken Q + text A) & \cmark &  &  & \cmark &  & \cmark \\
Multiple Choice (text Q) & \cmark &  & \cmark &  &  &  \\
Open-Ended Questions (text) & \cmark &  & \cmark &  &  & \cmark \\
Acoustic Consistency (AC) &  &  & \cmark & \cmark & \cmark &  \\
Acoustic–Semantic Alignment (ASA) &  &  & \cmark & \cmark & \cmark &  \\
Subjective MOS &  &  &  &  &  & \cmark \\
DNN-based MOS &  &  &  & \cmark &  &  \\
NISQA &  &  &  & \cmark &  &  \\
Meta AudioBox Aesthetics &  &  & \cmark & \cmark &  &  \\
ABX\_POS / ABX\_SEM & \cmark &  & \cmark &  &  &  \\
S2S Arena & \cmark &  &  &  &  & \cmark \\
Pause Handling &  &  & \cmark &  & \cmark &  \\
Backchanneling &  &  & \cmark &  & \cmark &  \\
Turn Taking &  &  & \cmark &  & \cmark &  \\
User Interruption Handling &  &  & \cmark & \cmark & \cmark &  \\

\end{supertabular}
\end{center}
\end{landscape}

%%%%%%%%%%%%%%%%%%%%%%%%%%%%%%%%%%%%%%%%%%%%%%%%%%%%%
% TABLE 5: BENCHMARK REFERENCE TABLE
%%%%%%%%%%%%%%%%%%%%%%%%%%%%%%%%%%%%%%%%%%%%%%%%%%%%%

\clearpage
\begin{landscape}
\begin{center}

\captionsetup{type=table}
\caption{Catalogue of major benchmarks mentioned in the taxonomy, including their applicability, included tasks, and descriptive context.}
\label{tab:benchmarks}

\tablefirsthead{
\toprule
\textbf{Name} & \textbf{Reference} & \textbf{Applicability} &
\textbf{Included Tasks} & \textbf{Languages} & \textbf{Short Description} \\
\midrule
}

\tablehead{
\multicolumn{6}{l}{\textbf{Table~\ref{tab:benchmarks} (continued): Benchmark catalogue}}\\
\toprule
\textbf{Name} & \textbf{Reference} & \textbf{Applicability} &
\textbf{Included Tasks} & \textbf{Languages} & \textbf{Short Description (continued)} \\
\midrule
}

\tabletail{
\midrule
\multicolumn{6}{r}{\textit{Table continued on next page}}\\
\midrule
}

\tablelasttail{\bottomrule}

\begin{supertabular}{@{}p{2.8cm} p{3.0cm} p{2.2cm} p{3.0cm} p{2.2cm} p{7.8cm}@{}}

SUPERB & \cite{yang2024large} & Speech &
PR; ASR; OOD-ASR; KS; QbE; SI; SV; SD; ER; IC; SF; ST; VC; SS; SE; LID &
English &
General benchmark for evaluating frozen speech foundation models on supervised downstream tasks. Models are fine-tuned with lightweight prediction heads; representations from all layers are linearly combined to form the final embedding. \\

ML-SUPERB & \cite{shi2023ml} & Speech &
ASR; LID; Joint Multilingual ASR/LID task &
Multilingual &
Multilingual extension of SUPERB covering cross-lingual and multilingual tasks. \\

SLUE / SLUE Phase-2 & \cite{shon2022slue,shon2023slue}  & Speech &
ASR; ER; DAC; QA; SUMM; NEL; NER &
English &
Benchmarks for speech language understanding tasks; tests models on sentiment, dialogue acts, summarization, and question answering using frozen encoders with task-specific classifiers. \\

FLEURS & \cite{conneau2023fleurs} & Speech &
ASR; LID; Cross-modal Retrieval &
Multilingual &
Multilingual dataset covering over 100 languages for ASR and cross-modal speech–text retrieval. \\

XTREME-S & \cite{conneau2022xtreme} & Speech &
ASR; ST; LID; IC; Speech–Text Retrieval &
Multilingual &
Multilingual suite extending XTREME to speech; tests ASR, translation, and cross-modal retrieval. \\

LeBenchmark / LeBenchmark 2.0 & \cite{evain2021lebenchmark,parcollet2024lebenchmark} & Speech &
ASR; CP; AST; ER; SV; POS; LAS; UAS &
French &
French speech benchmark combining recognition, semantics, and paralinguistic tasks; focuses on representation learning and fine-tuning for domain adaptation. \\

Codec-SUPERB & \cite{wu2024codec} & Speech &
PESQ; STOI; STFTDistance; MelDistance; F0CORR; AEC\_regen; ER\_regen; SV\_regen; ASR\_regen &
— &
Evaluation suite for assessing resynthesis fidelity of codec and generative models using both perceptual and task-based metrics. \\

NOSS & \cite{shor2020towards} & Speech &
SI; ER; LID &
English &
Benchmark for non-semantic speech tasks; measures paralinguistic and indexical information content in representations. \\

Zero Resource Speech Benchmark & \cite{ versteegh2015zero, dunbar2017zero, dunbar2019zero, nguyen2020zero} & Speech &
ABX; sWUGGY; sBLIMP; ProsAudit; sSIMI &
English; Multilingual (partial) &
Evaluates self-supervised and unsupervised speech representations on phonetic, prosodic, and syntactic discrimination tasks without labels. \\

MAD Speech & \cite{futeral2025mad} & Speech &
Voice / Gender / Emotion / Accent / Noise Diversity &
English &
Evaluates acoustic diversity in generated speech using facet-specific projection models; quantifies diversity with cosine dissimilarity and Vendi Scores, correlated with reference diversity statistics. \\

MMAU & S\cite{sakshi2024mmau} & Multimodal (Speech–Text) &
MCQ (text) &
— &
Multimodal benchmark testing reasoning, cultural understanding, and scene interpretation from audio-text pairs using multiple-choice questions. \\

VoxEval & \cite{cui2025voxeval} & Speech &
MCQ (spoken input + output) &
English &
Tests reasoning and knowledge understanding from spoken inputs and outputs; focuses on mathematical and factual reasoning derived from MMLU. \\

VoiceBench & \cite{chen2024voicebench} & Speech &
MCQ (spoken input + text output); Refusal tests &
English &
Evaluates reasoning and fairness in spoken models using both spoken and textual answers. \\

AudioBench & \cite{wang2025audiobench} & Multimodal (Speech–Text) &
Open-ended Questions; Answer Analysis &
— &
Instruction-following benchmark for audio LLMs; evaluates textual or speech outputs with LLM-as-judge scoring. \\

AirBench & \cite{yang2024air} & Multimodal (Speech–Text) &
Open-ended Questions; MCQ (text) &
— &
Combines foundational and conversational evaluation for audio LLMs, using GPT-4 to generate and judge answers to speech-based prompts. \\

SALMon & \cite{maimon2025salmon} & Speech / Acoustic &
AC; ASA &
English &
Evaluates acoustic consistency and acoustic–semantic alignment by comparing model probabilities for matched vs. mismatched acoustic configurations. \\

EmphAssess & \cite{seyssel2024emphassess} & Speech &
EmphAssess &
— &
Tests preservation of emphasis in regenerated or translated speech; compares semantic alignment and prosodic prominence. \\

StressTest & \cite{yosha2025stresstest} & Speech &
Sentence Stress Detection; Sentence Stress Reasoning &
English &
Benchmark for prosody-based reasoning and stress perception in spoken language models; 
evaluates models’ ability to identify and interpret prosodic emphasis. \\

Meta AudioBox & \cite{tjandra2025meta} & Acoustic / Generation &
AudioBox Aesthetics &
— &
Assesses aesthetic and production quality of generated audio using trained aesthetic predictors. Scores along four axes: Production Quality, Production Complexity, Content Enjoyment, and Content Usefulness. \\

Dynamic-SUPERB & \cite{huang2024dynamic} & Multimodal (Speech–Text) &
MCQ (text) &
— &
Instruction-tuned variant of SUPERB extending evaluation to multimodal and generative setups with textual and speech outputs. \\

ParaLBench & \cite{zhang2024paralbench} & Speech &
Emotion; Sentiment; Sarcasm; Gender; Age; Accent; Dialect &
— &
Evaluates paralinguistic and affective tasks using frozen foundation model embeddings and lightweight classification heads. \\

S2S Arena & \cite{jiang2025s2s} & Speech &
S2S Arena &
— &
Evaluates instruction-following capabilities of speech-to-speech models; requires pairwise human preference judgments. \\

Full-Duplex-Bench & \cite{lin2025full} & Speech &
Pause Handling; Backchanneling; Turn-Taking; Interruption Handling &
English &
Evaluates full-duplex spoken dialogue models on real-time coordination behaviours (pauses, overlaps, interruptions). Uses automatic metrics (e.g., takeover rate, latency) from external ASR tools such as CrispyWhisper. \\

\end{supertabular}
\end{center}
\end{landscape}

\bibliographystyle{elsarticle-num}
\bibliography{custom}

\begin{thebibliography}{10}
\expandafter\ifx\csname url\endcsname\relax
  \def\url#1{\texttt{#1}}\fi
\expandafter\ifx\csname urlprefix\endcsname\relax\def\urlprefix{URL }\fi
\expandafter\ifx\csname href\endcsname\relax
  \def\href#1#2{#2} \def\path#1{#1}\fi

\bibitem{yang2024large}
S.-w. Yang, H.-J. Chang, Z.~Huang, A.~T. Liu, C.-I. Lai, H.~Wu, J.~Shi, X.~Chang, H.-S. Tsai, W.-C. Huang, et~al., A large-scale evaluation of speech foundation models, IEEE/ACM Transactions on Audio, Speech, and Language Processing (2024).

\bibitem{shon2022slue}
S.~Shon, A.~Pasad, F.~Wu, P.~Brusco, Y.~Artzi, K.~Livescu, K.~J. Han, Slue: New benchmark tasks for spoken language understanding evaluation on natural speech, in: ICASSP 2022-2022 IEEE International Conference on Acoustics, Speech and Signal Processing (ICASSP), IEEE, 2022, pp. 7927--7931.

\bibitem{shon2023slue}
S.~Shon, S.~Arora, C.-J. Lin, A.~Pasad, F.~Wu, R.~Sharma, W.-L. Wu, H.-Y. Lee, K.~Livescu, S.~Watanabe, Slue phase-2: A benchmark suite of diverse spoken language understanding tasks, in: Proceedings of the 61st Annual Meeting of the Association for Computational Linguistics (Volume 1: Long Papers), 2023, pp. 8906--8937.

\bibitem{evain2021lebenchmark}
S.~Evain, H.~Nguyen, H.~Le, M.~Z. Boito, S.~Mdhaffar, S.~Alisamir, Z.~Tong, N.~Tomashenko, M.~Dinarelli, T.~Parcollet, et~al., Lebenchmark: A reproducible framework for assessing self-supervised representation learning from speech, in: INTERSPEECH 2021: Conference of the International Speech Communication Association, 2021.

\bibitem{parcollet2024lebenchmark}
T.~Parcollet, H.~Nguyen, S.~Evain, M.~Z. Boito, A.~Pupier, S.~Mdhaffar, H.~Le, S.~Alisamir, N.~Tomashenko, M.~Dinarelli, et~al., Lebenchmark 2.0: A standardized, replicable and enhanced framework for self-supervised representations of french speech, Computer Speech \& Language 86 (2024) 101622.

\bibitem{dunbar2021zero}
E.~Dunbar, M.~Bernard, N.~Hamilakis, T.~A. Nguyen, M.~de~Seyssel, P.~Roz{\'e}, M.~Rivi{\`e}re, E.~Kharitonov, E.~Dupoux, The zero resource speech challenge 2021: Spoken language modelling, Interspeech 2021 (2021).

\bibitem{wu2024codec}
H.~Wu, H.-L. Chung, Y.-C. Lin, Y.-K. Wu, X.~Chen, Y.-C. Pai, H.-H. Wang, K.-W. Chang, A.~Liu, H.-Y. Lee, Codec-superb: An in-depth analysis of sound codec models, in: Findings of the Association for Computational Linguistics ACL 2024, 2024, pp. 10330--10348.

\bibitem{maimon2025salmon}
G.~Maimon, A.~Roth, Y.~Adi, Salmon: A suite for acoustic language model evaluation, in: 2025 IEEE International Conference on Acoustics, Speech, and Signal Processing, ICASSP 2025, Institute of Electrical and Electronics Engineers, 2025.

\bibitem{Bommasani2021FoundationModels}
R.~Bommasani, D.~A. Hudson, E.~Adeli, R.~Altman, S.~Arora, S.~von Arx, M.~S. Bernstein, J.~Bohg, A.~Bosselut, E.~Brunskill, E.~Brynjolfsson, S.~Buch, D.~Card, R.~Castellon, N.~S. Chatterji, A.~S. Chen, K.~A. Creel, J.~Davis, D.~Demszky, C.~Donahue, M.~Doumbouya, E.~Durmus, S.~Ermon, J.~Etchemendy, K.~Ethayarajh, L.~Fei-Fei, C.~Finn, T.~Gale, L.~E. Gillespie, K.~Goel, N.~D. Goodman, S.~Grossman, N.~Guha, T.~Hashimoto, P.~Henderson, J.~Hewitt, D.~E. Ho, J.~Hong, K.~Hsu, J.~Huang, T.~F. Icard, S.~Jain, D.~Jurafsky, P.~Kalluri, S.~Karamcheti, G.~Keeling, F.~Khani, O.~Khattab, P.~W. Koh, M.~S. Krass, R.~Krishna, R.~Kuditipudi, A.~Kumar, F.~Ladhak, M.~Lee, T.~Lee, J.~Leskovec, I.~Levent, X.~L. Li, X.~Li, T.~Ma, A.~Malik, C.~D. Manning, S.~P. Mirchandani, E.~Mitchell, Z.~Munyikwa, S.~Nair, A.~Narayan, D.~Narayanan, B.~Newman, A.~Nie, J.~C. Niebles, H.~Nilforoshan, J.~F. Nyarko, G.~Ogut, L.~Orr, I.~Papadimitriou, J.~S. Park, C.~Piech, E.~Portelance, C.~Potts, A.~Raghunathan, R.~Reich, H.~Ren, F.~Rong, Y.~H. Roohani,
  C.~Ruiz, J.~Ryan, C.~R'e, D.~Sadigh, S.~Sagawa, K.~Santhanam, A.~Shih, K.~P. Srinivasan, A.~Tamkin, R.~Taori, A.~W. Thomas, F.~Tram{\`e}r, R.~E. Wang, W.~Wang, B.~Wu, J.~Wu, Y.~Wu, S.~M. Xie, M.~Yasunaga, J.~You, M.~A. Zaharia, M.~Zhang, T.~Zhang, X.~Zhang, Y.~Zhang, L.~Zheng, K.~Zhou, P.~Liang, \href{https://crfm.stanford.edu/assets/report.pdf}{On the opportunities and risks of foundation models}, ArXiv (2021).
\newline\urlprefix\url{https://crfm.stanford.edu/assets/report.pdf}

\bibitem{baevski2020wav2vec}
A.~Baevski, Y.~Zhou, A.~Mohamed, M.~Auli, wav2vec 2.0: A framework for self-supervised learning of speech representations, Advances in neural information processing systems 33 (2020) 12449--12460.

\bibitem{hsu2021hubert}
W.-N. Hsu, B.~Bolte, Y.-H.~H. Tsai, K.~Lakhotia, R.~Salakhutdinov, A.~Mohamed, Hubert: Self-supervised speech representation learning by masked prediction of hidden units, IEEE/ACM Transactions on Audio, Speech, and Language Processing 29 (2021) 3451--3460.

\bibitem{chen2022wavlm}
S.~Chen, C.~Wang, Z.~Chen, Y.~Wu, S.~Liu, Z.~Chen, J.~Li, N.~Kanda, T.~Yoshioka, X.~Xiao, et~al., Wavlm: Large-scale self-supervised pre-training for full stack speech processing, IEEE Journal of Selected Topics in Signal Processing 16~(6) (2022) 1505--1518.

\bibitem{mohamed2022self}
A.~Mohamed, H.-y. Lee, L.~Borgholt, J.~D. Havtorn, J.~Edin, C.~Igel, K.~Kirchhoff, S.-W. Li, K.~Livescu, L.~Maal{\o}e, et~al., Self-supervised speech representation learning: A review, IEEE Journal of Selected Topics in Signal Processing 16~(6) (2022) 1179--1210.

\bibitem{arora2025landscape}
S.~Arora, K.-W. Chang, C.-M. Chien, Y.~Peng, H.~Wu, Y.~Adi, E.~Dupoux, H.-Y. Lee, K.~Livescu, S.~Watanabe, On the landscape of spoken language models: A comprehensive survey, arXiv preprint arXiv:2504.08528 (2025).

\bibitem{latif2023sparks}
S.~Latif, M.~Shoukat, F.~Shamshad, M.~Usama, Y.~Ren, H.~Cuay{\'a}huitl, W.~Wang, X.~Zhang, R.~Togneri, E.~Cambria, et~al., Sparks of large audio models: A survey and outlook, arXiv preprint arXiv:2308.12792 (2023).

\bibitem{shi2023ml}
J.~Shi, D.~Berrebbi, W.~Chen, E.-P. Hu, W.-P. Huang, H.-L. Chung, X.~Chang, S.-W. Li, A.~Mohamed, H.-y. Lee, et~al., Ml-superb: Multilingual speech universal performance benchmark, in: Proc. Interspeech 2023, 2023, pp. 884--888.

\bibitem{conneau2023fleurs}
A.~Conneau, M.~Ma, S.~Khanuja, Y.~Zhang, V.~Axelrod, S.~Dalmia, J.~Riesa, C.~Rivera, A.~Bapna, Fleurs: Few-shot learning evaluation of universal representations of speech, in: 2022 IEEE Spoken Language Technology Workshop (SLT), IEEE, 2023, pp. 798--805.

\bibitem{conneau2022xtreme}
A.~Conneau, A.~Bapna, Y.~Zhang, M.~Ma, P.~von Platen, A.~Lozhkov, C.~Cherry, Y.~Jia, C.~Rivera, M.~Kale, et~al., Xtreme-s: Evaluating cross-lingual speech representations, in: Proc. Interspeech 2022, 2022, pp. 3248--3252.

\bibitem{versteegh2015zero}
M.~Versteegh, R.~Thiolliere, T.~Schatz, X.~Anguera, A.~Jansen, E.~Dupoux, The zero resource speech challenge 2015., ISCA, 2015, pp. 3169--3173.

\bibitem{dunbar2017zero}
E.~Dunbar, X.~N. Cao, J.~Benjumea, J.~Karadayi, M.~Bernard, L.~Besacier, X.~Anguera, E.~Dupoux, The zero resource speech challenge 2017, in: 2017 IEEE Automatic Speech Recognition and Understanding Workshop (ASRU), 2017, pp. 323--330.
\newblock \href {https://doi.org/10.1109/ASRU.2017.8268953} {\path{doi:10.1109/ASRU.2017.8268953}}.

\bibitem{dunbar2019zero}
E.~Dunbar, R.~Algayres, J.~Karadayi, M.~Bernard, J.~Benjumea, X.-N. Cao, L.~Miskic, C.~Dugrain, L.~Ondel, A.~W. Black, et~al., The zero resource speech challenge 2019: Tts without t, Interspeech 2019 (2019) 1088--1092.

\bibitem{nguyen2020zero}
T.~A. Nguyen, M.~de~Seyssel, P.~Roz{\'e}, M.~Rivi{\`e}re, E.~Kharitonov, A.~Baevski, E.~Dunbar, E.~Dupoux, The zero resource speech benchmark 2021: Metrics and baselines for unsupervised spoken language modeling, in: NeuRIPS Workshop on Self-Supervised Learning for Speech and Audio Processing, 2020.

\bibitem{futeral2025mad}
M.~Futeral, A.~Agostinelli, M.~Tagliasacchi, N.~Zeghidour, E.~Kharitonov, Mad speech: Measures of acoustic diversity of speech, in: Proceedings of the 2025 Conference of the Nations of the Americas Chapter of the Association for Computational Linguistics: Human Language Technologies (Volume 1: Long Papers), 2025, pp. 222--235.

\bibitem{seyssel2024emphassess}
M.~Seyssel, A.~D’Avirro, A.~Williams, E.~Dupoux, Emphassess: a prosodic benchmark on assessing emphasis transfer in speech-to-speech models, in: Proceedings of the 2024 Conference on Empirical Methods in Natural Language Processing, 2024, pp. 495--507.

\bibitem{yosha2025stresstest}
I.~Yosha, G.~Maimon, Y.~Adi, Stresstest: Can your speech lm handle the stress?, arXiv preprint arXiv:2505.22765 (2025).

\bibitem{lin2025full}
G.-T. Lin, J.~Lian, T.~Li, Q.~Wang, G.~Anumanchipalli, A.~H. Liu, H.-y. Lee, Full-duplex-bench: A benchmark to evaluate full-duplex spoken dialogue models on turn-taking capabilities, arXiv preprint arXiv:2503.04721 (2025).

\bibitem{cui2024recent}
W.~Cui, D.~Yu, X.~Jiao, Z.~Meng, G.~Zhang, Q.~Wang, Y.~Guo, I.~King, Recent advances in speech language models: A survey, arXiv preprint arXiv:2410.03751 (2024).

\bibitem{yang2025towards}
C.-K. Yang, N.~S. Ho, H.-y. Lee, Towards holistic evaluation of large audio-language models: A comprehensive survey, arXiv preprint arXiv:2505.15957 (2025).

\bibitem{schatz2013evaluating}
T.~Schatz, V.~Peddinti, F.~Bach, A.~Jansen, H.~Hermansky, E.~Dupoux, Evaluating speech features with the minimal-pair abx task: Analysis of the classical mfc/plp pipeline, in: Proceedings of Interspeech, 2013, pp. 1--5.

\bibitem{schatz2014evaluating}
T.~Schatz, V.~Peddinti, X.-N. Cao, F.~R. Bach, H.~Hermansky, E.~Dupoux, Evaluating speech features with the minimal-pair abx task (ii): resistance to noise., in: Proceedings of Interspeech, 2014, pp. 915--919.

\bibitem{de2023prosaudit}
M.~de~Seyssel, M.~Lavechin, H.~Titeux, A.~Thomas, G.~Virlet, A.~S. Revilla, G.~Wisniewski, B.~Ludusan, E.~Dupoux, Prosaudit, a prosodic benchmark for self-supervised speech models, in: Proc. Interspeech 2023, 2023, pp. 2963--2967.

\bibitem{saeki2024speechbertscore}
T.~Saeki, S.~Maiti, S.~Takamichi, S.~Watanabe, H.~Saruwatari, Speechbertscore: Reference-aware automatic evaluation of speech generation leveraging nlp evaluation metrics, in: Proc. Interspeech 2024, 2024, pp. 4943--4947.

\bibitem{maiti2023speechlmscore}
S.~Maiti, Y.~Peng, T.~Saeki, S.~Watanabe, Speechlmscore: Evaluating speech generation using speech language model, in: ICASSP 2023-2023 IEEE International Conference on Acoustics, Speech and Signal Processing (ICASSP), IEEE, 2023, pp. 1--5.

\bibitem{lakhotia2021generative}
K.~Lakhotia, E.~Kharitonov, W.-N. Hsu, Y.~Adi, A.~Polyak, B.~Bolte, T.-A. Nguyen, J.~Copet, A.~Baevski, A.~Mohamed, et~al., On generative spoken language modeling from raw audio, Transactions of the Association for Computational Linguistics 9 (2021) 1336--1354.

\bibitem{hassid2023textually}
M.~Hassid, T.~Remez, T.~A. Nguyen, I.~Gat, A.~Conneau, F.~Kreuk, J.~Copet, A.~Defossez, G.~Synnaeve, E.~Dupoux, et~al., Textually pretrained speech language models, Advances in Neural Information Processing Systems 36 (2023) 63483--63501.

\bibitem{sakshi2024mmau}
S.~Sakshi, U.~Tyagi, S.~Kumar, A.~Seth, R.~Selvakumar, O.~Nieto, R.~Duraiswami, S.~Ghosh, D.~Manocha, Mmau: A massive multi-task audio understanding and reasoning benchmark, arXiv preprint arXiv:2410.19168 (2024).

\bibitem{cui2025voxeval}
W.~Cui, X.~Jiao, Z.~Meng, I.~King, Voxeval: Benchmarking the knowledge understanding capabilities of end-to-end spoken language models, arXiv preprint arXiv:2501.04962 (2025).

\bibitem{wang2025audiobench}
B.~Wang, X.~Zou, G.~Lin, S.~Sun, Z.~Liu, W.~Zhang, Z.~Liu, A.~Aw, N.~Chen, Audiobench: A universal benchmark for audio large language models, in: Proceedings of the 2025 Conference of the Nations of the Americas Chapter of the Association for Computational Linguistics: Human Language Technologies (Volume 1: Long Papers), 2025, pp. 4297--4316.

\bibitem{yang2024air}
Q.~Yang, J.~Xu, W.~Liu, Y.~Chu, Z.~Jiang, X.~Zhou, Y.~Leng, Y.~Lv, Z.~Zhao, C.~Zhou, et~al., Air-bench: Benchmarking large audio-language models via generative comprehension, in: Proceedings of the 62nd Annual Meeting of the Association for Computational Linguistics (Volume 1: Long Papers), 2024, pp. 1979--1998.

\bibitem{chen2024voicebench}
Y.~Chen, X.~Yue, C.~Zhang, X.~Gao, R.~T. Tan, H.~Li, Voicebench: Benchmarking llm-based voice assistants, arXiv preprint arXiv:2410.17196 (2024).

\bibitem{shor2020towards}
J.~Shor, A.~Jansen, R.~Maor, O.~Lang, O.~Tuval, F.~d.~C. Quitry, M.~Tagliasacchi, I.~Shavitt, D.~Emanuel, Y.~Haviv, Towards learning a universal non-semantic representation of speech, in: Proc. Interspeech 2020, 2020, pp. 140--144.

\bibitem{jiang2025s2s}
F.~Jiang, Z.~Lin, F.~Bu, Y.~Du, B.~Wang, H.~Li, S2s-arena, evaluating speech2speech protocols on instruction following with paralinguistic information, arXiv preprint arXiv:2503.05085 (2025).

\bibitem{rix2001perceptual}
A.~W. Rix, J.~G. Beerends, M.~P. Hollier, A.~P. Hekstra, Perceptual evaluation of speech quality (pesq)-a new method for speech quality assessment of telephone networks and codecs, in: 2001 IEEE international conference on acoustics, speech, and signal processing. Proceedings (Cat. No. 01CH37221), Vol.~2, IEEE, 2001, pp. 749--752.

\bibitem{taal2011algorithm}
C.~H. Taal, R.~C. Hendriks, R.~Heusdens, J.~Jensen, An algorithm for intelligibility prediction of time--frequency weighted noisy speech, IEEE TRANSACTIONS ON AUDIO, SPEECH, AND LANGUAGE PROCESSING 19~(7) (2011) 2125.

\bibitem{vincent2006sdr}
E.~Vincent, R.~Gribonval, C.~Fevotte, Performance measurement in blind audio source separation, IEEE Transactions on Audio, Speech, and Language Processing 14~(4) (2006) 1462--1469.
\newblock \href {https://doi.org/10.1109/TSA.2005.858005} {\path{doi:10.1109/TSA.2005.858005}}.

\bibitem{kilgour2019frechet}
K.~Kilgour, M.~Zuluaga, D.~Roblek, M.~Sharifi, Fr{\'e}chet audio distance: A reference-free metric for evaluating music enhancement algorithms, in: Proc. Interspeech 2019, 2019, pp. 2350--2354.

\bibitem{mittag2019nisqa}
G.~Mittag, S.~Möller, Non-intrusive speech quality assessment for super-wideband speech communication networks, in: ICASSP 2019 - 2019 IEEE International Conference on Acoustics, Speech and Signal Processing (ICASSP), 2019, pp. 7125--7129.
\newblock \href {https://doi.org/10.1109/ICASSP.2019.8683770} {\path{doi:10.1109/ICASSP.2019.8683770}}.

\bibitem{reddy2021dnsmos}
C.~K. Reddy, V.~Gopal, R.~Cutler, Dnsmos: A non-intrusive perceptual objective speech quality metric to evaluate noise suppressors, in: ICASSP 2021-2021 IEEE International Conference on Acoustics, Speech and Signal Processing (ICASSP), IEEE, 2021, pp. 6493--6497.

\bibitem{tjandra2025meta}
A.~Tjandra, Y.-C. Wu, B.~Guo, J.~Hoffman, B.~Ellis, A.~Vyas, B.~Shi, S.~Chen, M.~Le, N.~Zacharov, et~al., Meta audiobox aesthetics: Unified automatic quality assessment for speech, music, and sound, arXiv preprint arXiv:2502.05139 (2025).

\bibitem{baevski2019vq}
A.~Baevski, S.~Schneider, M.~Auli, vq-wav2vec: Self-supervised learning of discrete speech representations, arXiv preprint arXiv:1910.05453 (2019).

\bibitem{defossez2022high}
A.~D{\'e}fossez, J.~Copet, G.~Synnaeve, Y.~Adi, High fidelity neural audio compression, arXiv preprint arXiv:2210.13438 (2022).

\bibitem{song2020score}
Y.~Song, J.~Sohl-Dickstein, D.~P. Kingma, A.~Kumar, S.~Ermon, B.~Poole, Score-based generative modeling through stochastic differential equations, arXiv preprint arXiv:2011.13456 (2020).

\bibitem{hendrycksmeasuring}
D.~Hendrycks, C.~Burns, S.~Basart, A.~Zou, M.~Mazeika, D.~Song, J.~Steinhardt, Measuring massive multitask language understanding, in: International Conference on Learning Representations, 2021.

\bibitem{kharitonov2022text}
E.~Kharitonov, A.~Lee, A.~Polyak, Y.~Adi, J.~Copet, K.~Lakhotia, T.-A. Nguyen, M.~Riviere, A.~Mohamed, E.~Dupoux, et~al., Text-free prosody-aware generative spoken language modeling, in: Proceedings of the 60th Annual Meeting of the Association for Computational Linguistics (Volume 1: Long Papers), 2022, pp. 8666--8681.

\bibitem{nguyen2025spirit}
T.~A. Nguyen, B.~Muller, B.~Yu, M.~R. Costa-Jussa, M.~Elbayad, S.~Popuri, C.~Ropers, P.-A. Duquenne, R.~Algayres, R.~Mavlyutov, et~al., Spirit-lm: Interleaved spoken and written language model, Transactions of the Association for Computational Linguistics 13 (2025) 30--52.

\bibitem{chu2024qwen2}
Y.~Chu, J.~Xu, Q.~Yang, H.~Wei, X.~Wei, Z.~Guo, Y.~Leng, Y.~Lv, J.~He, J.~Lin, et~al., Qwen2-audio technical report, arXiv preprint arXiv:2407.10759 (2024).

\bibitem{fang2024llama}
Q.~Fang, S.~Guo, Y.~Zhou, Z.~Ma, S.~Zhang, Y.~Feng, Llama-omni: Seamless speech interaction with large language models, arXiv preprint arXiv:2409.06666 (2024).

\bibitem{nguyen2023generative}
T.~A. Nguyen, E.~Kharitonov, J.~Copet, Y.~Adi, W.-N. Hsu, A.~Elkahky, P.~Tomasello, R.~Algayres, B.~Sagot, A.~Mohamed, et~al., Generative spoken dialogue language modeling, Transactions of the Association for Computational Linguistics 11 (2023) 250--266.

\bibitem{defossez2024moshi}
A.~D{\'e}fossez, L.~Mazar{\'e}, M.~Orsini, A.~Royer, P.~P{\'e}rez, H.~J{\'e}gou, E.~Grave, N.~Zeghidour, Moshi: a speech-text foundation model for real-time dialogue, arXiv preprint arXiv:2410.00037 (2024).

\bibitem{yu2024salmonn}
W.~Yu, S.~Wang, X.~Yang, X.~Chen, X.~Tian, J.~Zhang, G.~Sun, L.~Lu, Y.~Wang, C.~Zhang, Salmonn-omni: A codec-free llm for full-duplex speech understanding and generation, arXiv preprint arXiv:2411.18138 (2024).

\bibitem{papineni2002bleu}
K.~Papineni, S.~Roukos, T.~Ward, W.-J. Zhu, Bleu: a method for automatic evaluation of machine translation, in: Proceedings of the 40th annual meeting of the Association for Computational Linguistics, 2002, pp. 311--318.

\bibitem{zhang2024paralbench}
Z.~Zhang, W.~Xu, Z.~Dong, K.~Wang, Y.~Wu, J.~Peng, R.~Wang, D.-Y. Huang, Paralbench: a large-scale benchmark for computational paralinguistics over acoustic foundation models, IEEE Transactions on Affective Computing~(01) (2024) 1--17.

\bibitem{manku2025emergenttts}
R.~R. Manku, Y.~Tang, X.~Shi, M.~Li, A.~Smola, Emergenttts-eval: Evaluating tts models on complex prosodic, expressiveness, and linguistic challenges using model-as-a-judge, arXiv preprint arXiv:2505.23009 (2025).

\bibitem{minixhofer2025ttsds2}
C.~Minixhofer, O.~Klejch, P.~Bell, Ttsds2: Resources and benchmark for evaluating human-quality text to speech systems, arXiv preprint arXiv:2506.19441 (2025).

\bibitem{lin2004rouge}
C.-Y. Lin, Rouge: A package for automatic evaluation of summaries, in: Text summarization branches out, 2004, pp. 74--81.

\bibitem{banerjee2005meteor}
S.~Banerjee, A.~Lavie, Meteor: An automatic metric for mt evaluation with improved correlation with human judgments, in: Proceedings of the acl workshop on intrinsic and extrinsic evaluation measures for machine translation and/or summarization, 2005, pp. 65--72.

\bibitem{zhang2019bertscore}
T.~Zhang, V.~Kishore, F.~Wu, K.~Q. Weinberger, Y.~Artzi, Bertscore: Evaluating text generation with bert, arXiv preprint arXiv:1904.09675 (2019).

\bibitem{kubichek1993mel}
R.~Kubichek, Mel-cepstral distance measure for objective speech quality assessment, in: Proceedings of IEEE pacific rim conference on communications computers and signal processing, Vol.~1, IEEE, 1993, pp. 125--128.

\bibitem{le2019sdr}
J.~Le~Roux, S.~Wisdom, H.~Erdogan, J.~R. Hershey, Sdr--half-baked or well done?, in: ICASSP 2019-2019 IEEE International Conference on Acoustics, Speech and Signal Processing (ICASSP), IEEE, 2019, pp. 626--630.

\bibitem{luo2019conv}
Y.~Luo, N.~Mesgarani, Conv-tasnet: Surpassing ideal time--frequency magnitude masking for speech separation, IEEE/ACM transactions on audio, speech, and language processing 27~(8) (2019) 1256--1266.

\bibitem{beerends2013perceptual}
J.~G. Beerends, C.~Schmidmer, J.~Berger, M.~Obermann, R.~Ullmann, J.~Pomy, M.~Keyhl, Perceptual objective listening quality assessment (polqa), the third generation itu-t standard for end-to-end speech quality measurement part i—temporal alignment, journal of the audio engineering society 61~(6) (2013) 366--384.

\bibitem{hines2015visqol}
A.~Hines, J.~Skoglund, A.~C. Kokaram, N.~Harte, Visqol: an objective speech quality model, EURASIP Journal on Audio, Speech, and Music Processing 2015~(1) (2015) 13.

\bibitem{chinen2020visqol}
M.~Chinen, F.~S. Lim, J.~Skoglund, N.~Gureev, F.~O'Gorman, A.~Hines, Visqol v3: An open source production ready objective speech and audio metric, in: 2020 twelfth international conference on quality of multimedia experience (QoMEX), IEEE, 2020, pp. 1--6.

\bibitem{huang2024dynamic}
C.-y. Huang, K.-H. Lu, S.-H. Wang, C.-Y. Hsiao, C.-Y. Kuan, H.~Wu, S.~Arora, K.-W. Chang, J.~Shi, Y.~Peng, et~al., Dynamic-superb: Towards a dynamic, collaborative, and comprehensive instruction-tuning benchmark for speech, in: ICASSP 2024-2024 IEEE International Conference on Acoustics, Speech and Signal Processing (ICASSP), IEEE, 2024, pp. 12136--12140.

\bibitem{international1996methods}
I.~T. U. T.~S. Sector, Methods for subjective determination of transmission quality, International Telecommunication Union, 1996.

\bibitem{algayres2022dp}
R.~Algayres, T.~Ricoul, J.~Karadayi, H.~Lauren{\c{c}}on, S.~Zaiem, A.~Mohamed, B.~Sagot, E.~Dupoux, Dp-parse: Finding word boundaries from raw speech with an instance lexicon, Transactions of the Association for Computational Linguistics 10 (2022) 1051--1065.

\end{thebibliography}




% %% For authoryear reference style

% \bibitem[Lamport(1994)]{lamport94}
%   \textit{\LaTeX: a document preparation system},
%   Addison Wesley, Massachusetts,

% \end{thebibliography}
\end{document}